\documentclass[sigconf]{acmart}
\makeatletter
\def\@ACM@checkaffil{
    \if@ACM@instpresent\else
    \ClassWarningNoLine{\@classname}{No institution present for an affiliation}%
    \fi
    \if@ACM@citypresent\else
    \ClassWarningNoLine{\@classname}{No city present for an affiliation}%
    \fi
    \if@ACM@countrypresent\else
        \ClassWarningNoLine{\@classname}{No country present for an affiliation}%
    \fi
}
\makeatother

\usepackage{booktabs} 

\AtBeginDocument{%
  \providecommand\BibTeX{{%
    \normalfont B\kern-0.5em{\scshape i\kern-0.25em b}\kern-0.8em\TeX}}}

\usepackage{graphics}

\usepackage{hyperref} 
\usepackage[ruled]{algorithm2e} 
\usepackage{soul}
\usepackage{amsmath}

\SetAlFnt{\small}
\SetAlCapFnt{\small}
\SetAlCapNameFnt{\small}
\SetAlCapHSkip{0pt}

\usepackage[capitalize]{cleveref}
\crefname{section}{Sec.}{Secs.}
\Crefname{section}{Section}{Sections}
\Crefname{table}{Table}{Tables}
\crefname{table}{Tab.}{Tabs.}
\Crefname{Figure}{Figure}{Figures}
\crefname{figure}{Fig.}{Figs.}
\Crefname{Equation}{Equation}{Equations}
\crefname{equation}{Eq.}{Eqs.}

\newcommand{\vect}[1]{{{\bf{#1}}}}
\newcommand{\mat}[1]{{{\bf{#1}}}}

\copyrightyear{2023} 
\acmYear{2023} 
\setcopyright{acmlicensed}\acmConference[SIGGRAPH '23 Conference Proceedings]{Special Interest Group on Computer Graphics and Interactive Techniques Conference Conference Proceedings}{August 6--10, 2023}{Los Angeles, CA, USA}
\acmBooktitle{Special Interest Group on Computer Graphics and Interactive Techniques Conference Conference Proceedings (SIGGRAPH '23 Conference Proceedings), August 6--10, 2023, Los Angeles, CA, USA}
\acmPrice{15.00}
\acmDOI{10.1145/3588432.3591514}
\acmISBN{979-8-4007-0159-7/23/08}

\begin{document}
\title{RSMT: Real-time Stylized Motion Transition for Characters}
\author{Xiangjun Tang}
\orcid{0000-0001-7441-0086}
\affiliation{%
 \institution{Zhejiang University
 }
 \city{Hangzhou}
  \country{China}
 }
 \email{fcsx1tf@163.com}

\author{Linjun Wu}
\orcid{0000-0002-1988-0090}
\affiliation{%
 \institution{ Zhejiang University
 }
  \city{Hangzhou}
  \country{China}
 }
 \email{3190104528@zju.edu.cn}

\author{He Wang}
\orcid{0000-0002-2281-5679}
\affiliation{%
 \institution{University of Leeds}
  \city{Leeds}
  \country{United Kingdom}
 }
 \email{H.E.Wang@leeds.ac.uk}

\author{Bo Hu}
\orcid{0000-0002-6599-7249}
\affiliation{%
 \institution{Tencent Technology Co., Ltd.}
 \city{Shenzhen}
  \country{China}
}
 \email{corehu@tencent.com}

\author{Xu Gong}
\orcid{0000-0003-3900-2903}
\affiliation{%
 \institution{Tencent Technology Co., Ltd.}
  \city{Shenzhen}
  \country{China}
}
 \email{xugong@tencent.com}

\author{Yuchen Liao}
\orcid{0000-0002-9008-3609}
\affiliation{%
 \institution{Tencent Technology Co., Ltd.}
  \city{Shenzhen}
  \country{China}
}
 \email{bluecatliao@tencent.com}

\author{Songnan Li}
\orcid{0000-0001-6931-4129}
\affiliation{%
 \institution{Tencent Technology Co., Ltd.}
  \city{Shenzhen}
  \country{China}
}
 \email{sunnysnli@tencent.com}

\author{Qilong Kou}
\orcid{0000-0002-5222-7069}
\affiliation{%
 \institution{Tencent Technology Co., Ltd.}
  \city{Shenzhen}
  \country{China}
}
 \email{rambokou@tencent.com}

\author{Xiaogang Jin}
\authornote{Corresponding author}
\orcid{0000-0001-7339-2920}
\affiliation{%
 \institution{ Zhejiang University
 }
 \city{Hangzhou}
  \country{China}
}
 \email{jin@cad.zju.edu.cn}
\newcommand{\TXJ}[1]{{\textcolor{blue}{#1}}}
\newcommand{\Revision}[1]{{\textcolor{blue}{#1}}}
\renewcommand\shortauthors{Tang et al.}
\begin{abstract}

Styled online in-between motion generation has important application scenarios in computer animation and games. Its core challenge lies in the need to satisfy four critical requirements simultaneously: generation speed,  motion quality, style diversity, and synthesis controllability. While the first two challenges demand a delicate balance between simple fast models and learning capacity for generation quality, the latter two are rarely investigated together in existing methods, which largely focus on either control without style or uncontrolled stylized motions. To this end, we propose a Real-time Stylized Motion Transition method (RSMT) to achieve all aforementioned goals. Our method consists of two critical, independent components: a general motion manifold model and a style motion sampler. The former acts as a high-quality motion source and the latter synthesizes styled motions on the fly under control signals. Since both components can be trained separately on different datasets, our method provides great flexibility, requires less data, and generalizes well when no/few samples are available for unseen styles. Through exhaustive evaluation, our method proves to be fast, high-quality, versatile, and controllable. {The code and data are available at \href{https://github.com/yuyujunjun/RSMT-Realtime-Stylized-Motion-Transition}{https://github.com/yuyujunjun/RSMT-Realtime-Stylized-Motion-Transition.}}
\end{abstract}

%
%

\ccsdesc[500]{Computing methodologies~Motion capture}
\ccsdesc[300]{Computing methodologies~Motion transition}
\ccsdesc{Computing methodologies~Neural networks}
\ccsdesc[100]{Computing methodologies~Motion manifold}

%
%

\keywords{Animation, real-time, locomotion, motion manifold, conditional transitioning, in-betweening, deep learning}

\begin{teaserfigure}
\centering
  \includegraphics[width=\textwidth]{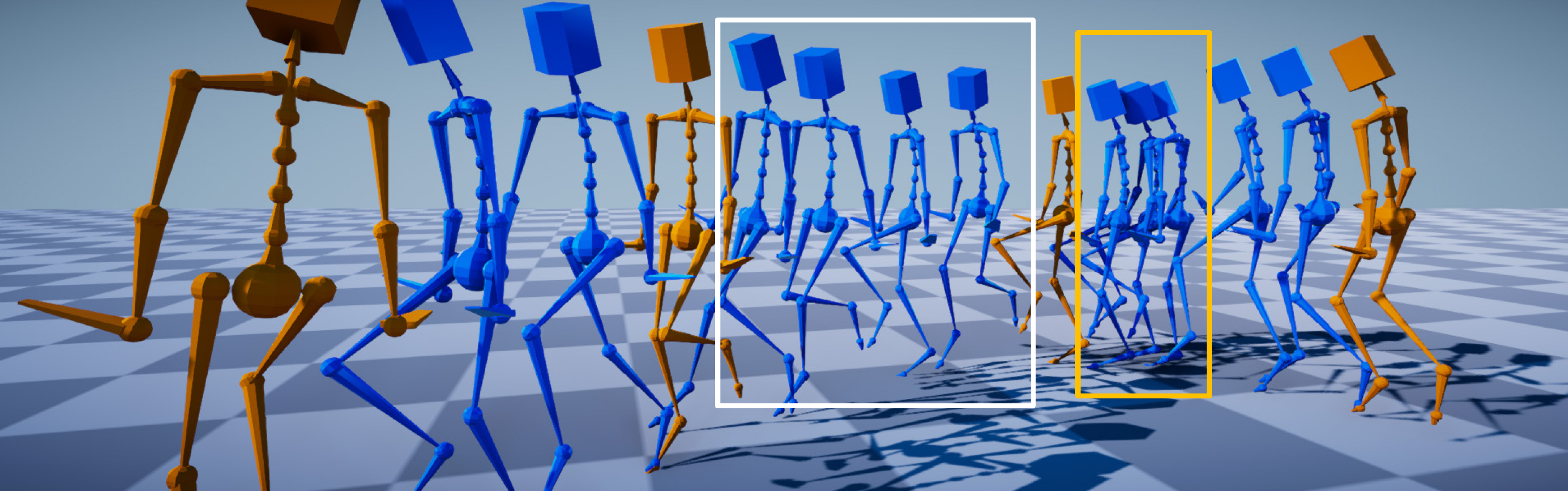}
  \caption{Our method generates an in-between motion sequence with a high leg lifting style (blue) between target frames (orange). Given a target frame, a desired transition duration and the required style, the character can dynamically adjust motions to reach the target with the desired style. Specifically, the character in the white box needs bigger steps to reach the target within the specified duration, while the character in the orange box needs smaller steps to reduce the speed for the same reason.}
  \Description{}
  \label{fig:teaser}
\end{teaserfigure}

\maketitle

\section{Introduction}

We investigate an under-explored motion generation problem: styled online in-between motion synthesis. 
{Rather than controlled style variation, our research focuses on maintaining a specific style while filling motions between keyframes, avoiding collapse to simple neutral motion, and has important applications in computer animation and games. }
It consists of two key elements that can potentially contradict with one another. Online in-between motion synthesis aims to quickly compute natural high-quality motions that satisfy the given constraints e.g. specified by frames~\cite{harvey2018recurrent,harvey_robust_2020}, while stylized motion generation focuses on motions that are visually different but with the same semantics (e.g. normal and zombie walking)~\cite{dong2020adult2child,yin2023dance}. Overall, the former attempts to find the optimal (in a broad sense) motion while the latter explores the motion diversity. 


Existing research mostly investigates the above two problems separately~\cite{qin2022motion,oreshkin2022delta,Mason2022Style,Mason_2018_fewshot}. Given the initial and end frame(s), in-between motions can be formulated into complex optimization problems~\cite{wang_energy_2015}. If data is available, it can be solved by searching in structured data~\cite{Kovar_motion_2008,Min_motion_2012,shen_posture_2017}, or maximizing the likelihood or a posteriori ~\cite{Li_Likelihood_estimation}. In deep learning, it can also be formulated into a motion manifold learning or control problem~\cite{Chen_dynamic_2020,holden_deep_2016,li2021task,petrovich2021action}. In parallel, motion style has also been separately investigated. Motion can be stylized via optimization~\cite{hsu2005style}, or learning parametric or non-parametric models on limited data~\cite{brand2000style,wang2007multifactor}. More recently, deep learning has been employed to extract and transfer the style features of motions~\cite{wen2021autoregressive,park2021diverse,du2019stylistic}.


Combining in-between and style motion needs to address several challenges simultaneously: generation speed, motion quality, style diversity, and synthesis controllability. The first two challenges dictate that simple models need to be employed and strictly rule out post-processing. Existing approaches are either too slow, rely on post-processing~\cite{duan_single-shot_2021,Jiang_motion_puzzle,aberman2020unpaired} or can only handle limited amount of data~\cite{chai_constraint-based_2007}. Further, existing research rarely considers the latter two challenges together, either focusing on control without style~\cite{tang2022cvae} or uncontrolled style motions~\cite{Mason_2018_fewshot,Mason2022Style}. Enforcing control can easily break the intended style and vice versa. 
\Cref{tab:highLevelComprison} shows the high level differences between our method and the literature.

\begin{table}[ht]
\vspace{-2.0em}
\caption{High level comparison between our method and existing methods.}
\label{tab:highLevelComprison}
\resizebox{1.0\columnwidth}{!}{
\begin{tabular}{llllll}
       & Speed & Quality & Style &  Controllability & large dataset\\ \hline
Optimization &$\times$  &$\times$ &$\checkmark$ & $\checkmark$ & n/a\\
Traditional in-between &$\checkmark$  &$\checkmark$ &$\times$ & $\checkmark$ & $\times$\\
Traditional style &$\checkmark$ &$\checkmark$ & $\checkmark$ &$\times$ & $\times$\\
Deep learning (in-between)&$\checkmark$ & $\checkmark$& $\times$&$\checkmark$&$\checkmark$\\
Deep learning (style)& $\checkmark$&$\checkmark$ & $\checkmark$ & $\times$&$\checkmark$\\  
Ours & $\checkmark$&$\checkmark$ &$\checkmark$ & $\checkmark$ &$\checkmark$
\end{tabular}
}
\end{table}

To this end, we propose a novel method, which can generate high-quality styled in-between motions in real time, given the starting frame, the end frame, the time duration, and a motion with the target style. 
Inspired by the work in motion manifold/representation learning~\cite{ling_character_2020,du2019stylistic} and conditional generation in images/motion~\cite{starke_neural_2021,huang2017arbitrary}, our key insight is to decouple the source of generation from the control of synthesis. Similar to representation learning, a good motion source should provide high-quality motions with varying styles. Furthermore, the extracted motion and style features should be ready for controlled synthesis, as a downstream task. This observation naturally leads to two key components in our system: a general motion manifold representation and a style motion sampler incorporating control. Moreover, such a design also allows the manifold and the sampler to be trained under different settings, e.g. the manifold trained on large unstyled/not-strongly-styled datasets widely available, and the sampler trained on smaller/dedicated datasets with desired styles, with little or no overlapping with the manifold data. This is ideal because the manifold can be pre-trained and shared, while the sampler is easily adaptable.

Our manifold model is an autoencoder that encodes motion transition randomness~\cite{tang2022cvae,ling_character_2020} and motion phases~\cite{Starke_deeppahse_2022}. The autoencoder encodes motions into a latent space conforming to a Gaussian distribution, which together with the extracted phase information jointly serve as a good source. A motion sampler based on recurrent neural networks is then employed to sample from the source, with specified target frames, time duration and style, avoiding generating only the most probable motions such as~\cite{tang2022cvae,harvey_robust_2020}. We exhaustively evaluate our method on 100STYLE dataset~\cite{Mason2022Style} and compare it with the most recent baseline methods~\cite{tang2022cvae,harvey_robust_2020} under multiple metrics regarding motion quality, controllability, style diversity, etc. Our method proves to be fast, high-quality, versatile, and controllable.


Our main contributions can be summarized as follows:

\begin{itemize}
    \item We present a novel online framework for styled real-time in-between motion generation without post-processing. 
    \item We propose a new method to combine styles and controllability in motion generation.
    \item We propose a new model that has strong generalization capacity in both motion control and stylization.
\end{itemize}

\section{Related work}
\subsection{In-between motion generation.}
In-between motion synthesis can be regarded as motion planning \cite{wang_energy_2011,arikan_interactive_2002, beaudoin2008motion, levine2012continuous,wang_harmonic_2013,Safonova07constructionand}, which solves complex optimization problems with various constraints. 
Data-driven methods are faster and produce more natural motions by searching in structured data, such as motion graphs \cite{Kovar_motion_2008,Min_motion_2012,shen_posture_2017}, but there is a trade-off between the diversity of animation and the space complexity. Recently, deep neural networks \cite{holden_learned_2020} leverage compressed data representations to reduce the space complexity. 

Learning-based online in-between motion synthesizing can be formulated as motion manifold learning problems~\cite{Chen_dynamic_2020,holden_deep_2016,li2021task,petrovich2021action,rempe2021humor,Wang_STRNN_2019,he2022nemf} with time-space constraints~\cite{harvey2018recurrent,harvey_robust_2020}.  
Besides, explicitly modeling the low-level motion transition as a motion manifold~\cite{tang2022cvae} improves the transition generalization for unseen control. However, without style control explicitly, the character reaches the target by the most probable motion, which might break the intended style.

Offline methods can be considered motion completion problems \cite{hernandez2019human,kaufmann_convolutional_2020}, which can be solved by convolution neural networks or transformers \cite{duan_single-shot_2021,qin2022motion,oreshkin2022delta}. Recent research employs generative models, such as diffusion models \cite{tevet2022diffusion} or generative adversarial networks \cite{Li_2022_GANimator} to fill the missing frame with the specific actions by user control. However, the models give little attention to time efficiency, so it's hard to employ them for real-time applications.

\subsection{Motion style transfer}
One solution to motion style transfer models the style variations between two motions \cite{hsu2005style}. Some research employs data-driven methods, such as Bayesian Networks \cite{ma2010modeling} or mixtures of autoregressive models \cite{xia2015realtime}. Unuma et al. \shortcite{unuma1995fourier} and  Yumer and Mitra \shortcite{yumer2016spectral} extracted style features between independent actions by transforming the motion into the frequency domain, but they largely handle relatively small amounts of data. Recent deep learning methods model the style directly. The style can be represented as one-hot embedding \cite{smith2019efficient}, where the style is stored as parameters of the neural network. However, one-hot embedding representation cannot be generalized to unseen styles. To solve this, Mason et al. \shortcite{Mason_2018_fewshot} proposed a residual block to model the style. Further, recent methods~\cite{aberman2020unpaired,Jiang_motion_puzzle} extract the style feature as the latent variable explicitly, making zero-shot learning available. When the content sequence is missing for in-between motion generation, the style transfer problem can be transformed into a style-guided generation problem. The most related work synthesizes motion sequences with the style feature introduced by the FiLM block~\cite{Mason2022Style}. However, generating the stylized in-between sequence that satisfies control signals is still an unsolved problem.

\section{Methodology}

A human motion with $T$ frames $\mat{M}=\{\vect{s}^{0\sim T}\}$ can be represented as a series of skeletal poses $\vect{s}^t \in \mathbb{R}^{N\times D}$, where $N$ is the joint number and $D$ is the degrees of freedom per joint at time $t\in[0, T]$. An often employed assumption on human motion is a linear dynamical system~\cite{li2002motion,pavlovic2000learning}:
\begin{equation}
\label{eq:LDS}
    s^{t+1} = s^t + f_1(h^{t+1}, s^t) \text{ and } h^{t+1} = f_2(h^t, s^t),
\end{equation}
where $h$ is a latent variable and governs the dynamics. When $f_1$ and $f_2$ are stochastic, we can model the joint probability of $M$ and $H=\{h^{0\sim T}\}$ as:
\begin{align}
    &P(M, H) = P(s^{1\sim T-1}, h^{1\sim T} | s^T, s^0, h^0,k)P(s^0)P(h^0)P(s^T)P(k) \nonumber \\
    &=P(s^0)P(h^0)P(s^T)P(k) \nonumber \\
    &\prod P(s^{t+1}|s^T, s^0, h^0, h^{t+1}, s^t)P(h_{t+1}|s^T, s^0, h^0, k, h^t, s^t),
\label{eq:PLDS}
\end{align}
where $s^0$, $h^0$ and $s^T$ can be seen as control variables for in-between motion generation and $k$ is a summarizing style code. The motion dynamics is governed by the last line in \cref{eq:PLDS}. Here we employ $h^t = \{z^t, v_h^t, p^t\}$, where $z^t$ is a latent variable to govern the stochasticity of the transition of $h$, $v_h^t$ is the hip velocity and orientation, $p^t$ is a phase-related feature. We omit $s^0$ and $h^0$ for simplicity in the rest of the paper. Learning can be conducted through maximizing \Cref{eq:PLDS} which is decomposed into two parts: the motion manifold $P(s^{t+1}|s^T, h^{t+1}, s^t)$ and the sampler $P(h_{t+1}|s^T, h^t, s^t,k)$. We model both by deep neural networks. {In summary, the sampler samples the stylized transition motion from the manifold by auto-regressively predicting the phase, hip, and latent variable.}
Below, we introduce their high-level design and refer the readers to the supplementary material for network details.

\subsection{The Motion Manifold}
Learning the motion manifold requires specifying the model for $P(s^{t+1}|s^T, h^{t+1}, s^t)$. First, instead of learning the next pose $s^{t+1}$ directly, we learn a pose change $\Delta s^{t+1} = s^{t+1} - s^t$, as given a fixed time interval, this is equivalent of learning a velocity profile which captures the dynamics better~\cite{martinez2017human}. Next, we drop $k$ and $s^T$ from $h^{t+1}$. This is because $k$ is a summarizing style code and $s^T$ is the target state, while we aim to make the manifold more focused on learning local natural motions with any style. So $P(s^{t+1}|s^T, h^{t+1}, s^t)$ = $P(\Delta s^{t+1}| v_h^{t+1}, p^{t+1}, z^{t+1}, s^t)$.
Directly learning it is not possible as only $\Delta s^{t+1}$, $v_h^{t+1}$ and $s^t$ are observable and $p^{t+1}$ and $z^{t+1}$ need to be learned as latent variables. Therefore, we propose to use an autoencoder to learn the mapping while regularizing $p^{t+1}$ and $z^{t+1}$. 

\begin{figure}[tb]
	\centering
	\includegraphics[width=1.05\linewidth]{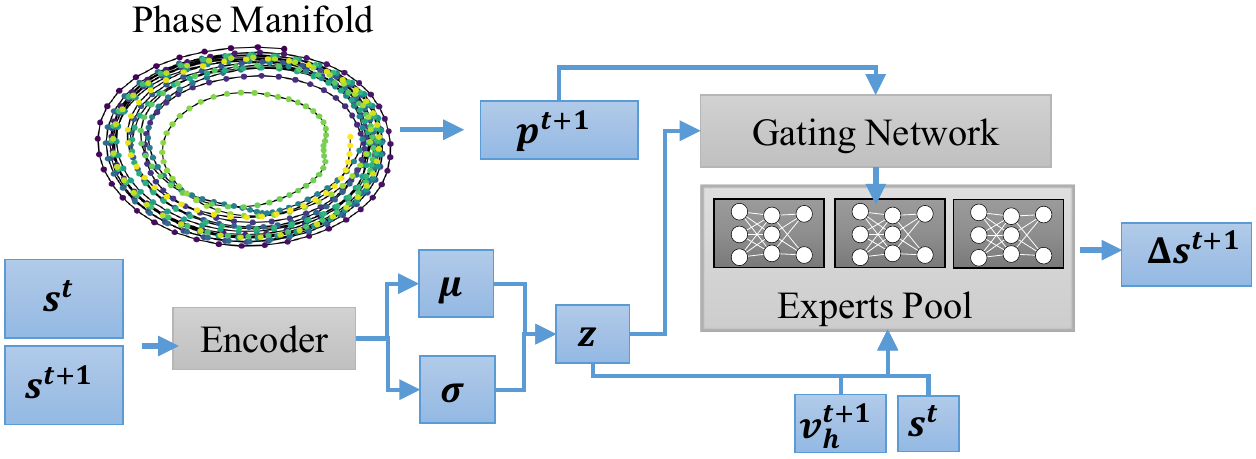}
    \caption{During training, the encoder takes $s^t$ and $s^{t+1}$ and generates the mean value $\mu$ and log variance $\sigma$ of the Gaussian distribution. Then the latent variable $z$ is calculated by reparameterization. Next, the gating network takes the phase sampled from the phase manifold and the latent variable $z$ to generate the blending coefficients for experts. Finally, the blended expert takes the current frame $s^t$, the future root $v_h^{t+1}$, and the latent vector $z$ to predict the pose change. }
	\label{fig:AE} 
 \vspace{-2em}
\end{figure}

The architecture of the autoencoder is shown in \cref{fig:AE}. The input is two consecutive frames and the output is the pose change. The encoder encodes the input into $z$ regularized by $z\sim N(0,1)$ to capture the transition stochasticity. The decoder recovers motions based on $z$, the style related phase features $p^{t+1}$, the current frame $s^t$, and the future hip velocity and orientation $v_h^{t+1}$. $v_h^{t+1}$ is a strong indicator of the next frame in terms of the general motion trend~\cite{tang2022cvae}, especially on the lower body where the motion diversity is smaller compared with the upper body. However, $v_h^{t+1}$ is insufficient to include style-related information, so we use another phase-related feature $p^{t+1}$. 

The phase feature in the latent space reveals well-behaved motion patterns~\cite{Starke_deeppahse_2022}. For our problem, it is complementary to $v_h^{t+1}$ in that it describes a continuous change of motion in patterns, which is highly related to styles~\cite{yumer2016spectral}. Following \cite{Starke_deeppahse_2022}, we train a periodic autoencoder to extract a multi-dimensional phase manifold. The resulting phase vectors on this manifold are represented by a signed phase shift $S$, which when multiplied by 2$\pi$ represents the angle of the phase vector, and an amplitude $A$, representing the phase vector magnitude:
\begin{equation}
    p_{2i-1} = A_i\cdot sin(2\pi \cdot S_i),\quad p_{2i} = A_i\cdot cos(2\pi \cdot S_i),
\end{equation}
where $i$ is the dimension index. After encoding, the decoder aims to recover the velocity, which is designed to be generative and learns $P(\Delta s^{t+1}|v_h^{t+1}, p^{t+1}, z^{t+1}, s^t)$. $z^{t+1}$ is drawn from $N(\mu, \sigma)$. $p^{t+1}$ is sampled from the phase manifold. Both are fed into a gating network for a Conditional Mixtures of Experts (CMoEs), conditioned on the current frame $s^t$, the future hip feature $v_h^{t+1}$ and $z^{t+1}$. The reason to employ a CMoEs here is that the transition distribution itself is multimodal~\cite{Wang_STRNN_2019,holden2017phase} and this multi-modality is further amplified with styles. A single network would average the data. In the CMoEs, each expert is expected to learn one mode and a combination of them can capture the multi-modality, with their weighting computed from $z^{t+1}$ and $p^{t+1}$. Here $z^{t+1}$ and $p^{t+1}$ also play different roles. $z^{t+1}$ governs the stochastic state transition of the linear dynamical system (\cref{eq:LDS}) while we find $p^{t+1}$ is a good continuous representation that encodes motion styles. This also means our motion manifold implicitly learns style-related features which can be utilized later by the sampler.

\subsubsection{Losses} We minimize $L = L_{rec} + \beta L_{kl} + L_{foot}$ where:

\begin{align}
\label{eq:loss_manifold}
    &L_{rec} = \frac{1}{ITND}\sum\lvert\lvert M_i - M'_i \rvert\rvert^2_2, \\
    &L_{kl} = -0.5\cdot (1+\sigma -\mu^2-e^{\sigma}) \text{,  } L_{foot} = \frac{1}{ITN_f}\sum\lvert\lvert f_v' \delta(f_v)\rvert\rvert^2_2, \nonumber
\end{align}
where $I$, $T$, $N$, $D$, and $N_f$ are the number of data samples, the motion length, the number of joints, the degrees of freedom per joint, and the number of foot joints that contact on the ground. $M_i$ and $M'_i$ are the ground-truth and predicted motion. Overall, $L_{rec}$ is a reconstruction loss. {$L_{foot}$ penalizes foot sliding, where $f_v'$ is the predicted foot velocity and $\delta(f_v) \in [0,1]$ is the probability of ground-truth foot contact for each foot in each frame,}
 which is defined by:
\begin{equation}
\delta(f_v) = 
\begin{cases} 1 ,& f_v\leq0.5,
\\
 0,&  f_v\geq1.0,\\
2t^3-3t^2+1,& t=2(f_v-0.5), f_v\in(0.5,1.0).
\end{cases}
\end{equation}
$L_{kl}$ is the KL-divergence between the $z$ distribution parameterized by $\mu$ and $\sigma$ and $N(0, 1)$, which we reparameterize following~\cite{kingma2013auto}. $\beta$ is a hyper-parameter balancing between quality and generalization, which is set to $0.001$~\cite{tang2022cvae,ling_character_2020}. Note we intentionally employ a relatively small $\beta$ to ensure the motion quality in the manifold. Later experiments will show that the generalization ability, which would be otherwise lost, will be compensated by the sampler.

\subsection{The Motion Sampler}
\begin{figure*}[h]
	\centering
	\includegraphics[width=0.9\linewidth]{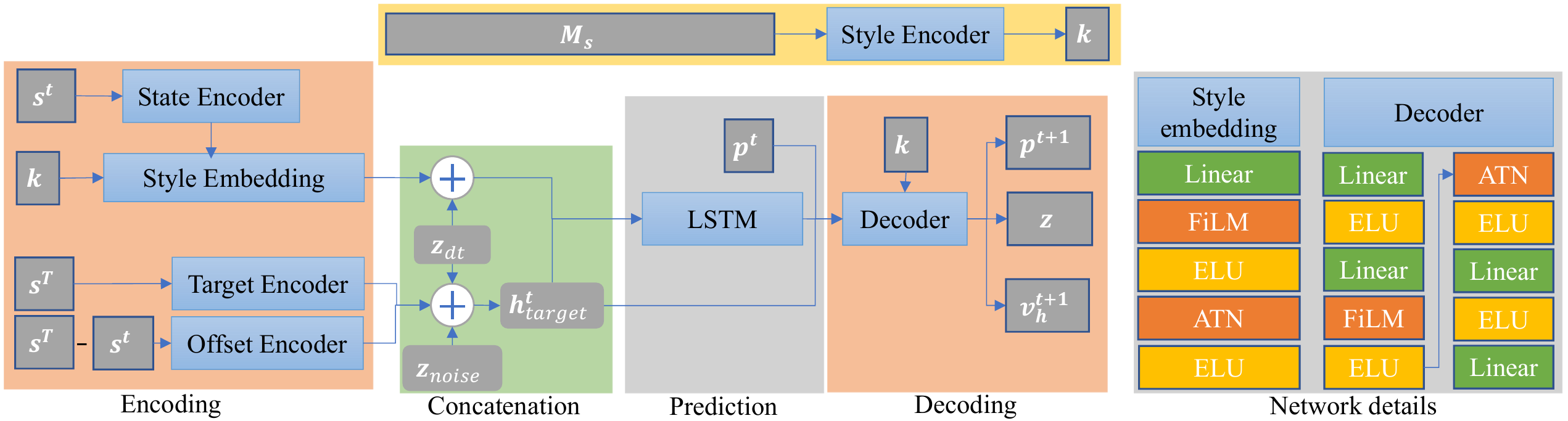}
    \caption{{{The motion sampler is divided into four stages: encoding, concatenation, prediction, and decoding}. The gray boxes represent the data and the blue ones represent network blocks. The merge of arrows represents vector concatenation.
    }}
	\label{fig:sampler}
 \vspace{-1em}
\end{figure*}
After training, the manifold is captured by the decoder and ready to be sampled. We learn the desired motions by learning the transition distribution $P(h_{t+1}|s^T, h^t, s^t)$. The sampler's design is shown in \cref{fig:sampler}. 
The sampler takes as input the current frame $s^t$, the target frame $s^T$, the current phase vector $p^t$ and the styled motion $M_s$ with the target style. 
Then it outputs the latent vector $h^{t+1} = \{p^{t+1}, z^{t+1}, v_h^{t+1}\}$ which can then be fed into the manifold motion to sample the next frame.

Specifically, we first extract a summarizing style code $k$ from $M_s$ by the style encoder based on convolutional neural networks. $k$ has a temporal dimension which we find is crucial in sampling, where the style information for each frame is lost in some earlier work~\cite{aberman2020unpaired,park2021diverse}. Next, $k$ is combined with the current frame $s^t$ in a style embedding after $s^t$ is pulled through a state encoder. This embedding co-embeds a single frame and a style code into a common space. In parallel, the target frame $s^{T}$ and the offset $s^{T} - s^t$ are individually encoded and concatenated, then perturbed by $z_{noise}$. This term captures the motion randomness in terms of completing the motion at the current time. The randomness is time-varying, i.e. small when $t$ is close to $T$ as the remaining motion is short, and bigger otherwise:
\begin{equation}
    z_{noise} = \lambda N(0, 0.5) 
    \text{ , }\lambda = clamp \left(\frac{T-t-t_{zero}}{t_{period}-t_{zero}}\right) \in (0, 1),
\end{equation}
where $t_{zero}=5$ and $t_{period}=30$ are parameters. So far, combining encoded information, we have:
{
\begin{align}
&z_h^{t+1} = \Phi_{LSTM}(c^t,  z_h^t), \nonumber \\
&c^t = (E_{sty}(k, E_{stat}(s^t))+ z_{dt}, h^t_{target}),  \nonumber \\
&h_{target}^t = (E_{tar}(s^T), E_{off}(s^T-s^t))+z_{noise} + z_{dt}, \nonumber \\
&z_{dt,2i} = {\rm sin} \left(\frac{dt}{10000^{2i/d}}\right),\quad z_{dt,2i+1}={\rm cos} \left(\frac{dt}{10000^{2i/d}}\right),
\end{align}
where $E_{sty}$, $E_{stat}$, $E_{tar}$, $E_{off}$, and $\Phi_{LSTM}$ are the style embedding, state encoder, target encoder, offset encoder, and long short-term memory network with cell state $c^t$, respectively.}
$d$ represents the dimension of $z_{dt}$ and $i\in [0,...,d/2]$ represents the dimension index. The $\Phi_{LSTM}$ predicts the next state $z_h^{t+1}$ which is used to predict $h^{t+1} = \{z^{t+1}, p^{t+1}, v_h^{t+1}\} = D(k, ( p^t, z_h^{t+1},h^t_{target}))$, which is further fed into the manifold model to sample the next frame. Instead of predicting the next phase directly~\cite{Starke_deeppahse_2022}, we predict an intermediate next phase ${\hat{p}^{t+1}}$, and the amplitude $\hat{A}^{t+1}$ and the frequency $\hat{F}^{t+1}$ separately. Then another intermediate phase vector is computed as:
\begin{align}
    \widetilde{p}^{t+1} = \hat{A}^{i+1}\cdot(R(\theta)\cdot p^t), \theta=\Delta t\cdot 2\pi \cdot \hat{F}^{t+1},
\end{align}
where $\Delta t$ is the time step, $R$ is a 2D rotation matrix. Then we interpolate the angles of $\widetilde{p}^{i+1}$ and $\hat{p}^{i+1}$ with spherical linear interpolation with weight $0.5$, and linearly interpolate the amplitude with weight $0.5$ for the final prediction $p^{t+1}$.

\subsubsection{Losses}
To train the sampler, we minimize $L = L_{rec}+L_{last}++ L_{foot}+L_{phase}$ where:

\begin{align}
\label{eq:transition_loss}
    L_{rec} =& \frac{1}{ITND}\sum ||M_i-M'_i||_1,   L_{last} = \frac{1}{IND}\sum ||s^T-(s^T)'||_1, \nonumber \\
    L_{phase} =& \frac{1}{ITN_{p}}\sum(||A^t-\hat{A}^t||_2^2+||F^t-\hat{F}^t||_2^2) \nonumber \\
    &+\frac{1}{2ITN_{p}}\sum(||p^t-\hat{p}^t||_2^2+||p^t-\widetilde{p}^t||_2^2),
\end{align}
where $I$, $T$, $N$, $D$, and $N_{p}$ are the number of data samples, the motion length, the number of joints, the degrees of freedom per joint, and the number of phase channels. 
$L_{rec}$ is the reconstruction loss of motion. $L_{last}$ emphasizes the predicted last frame should be the same as the target frame. $L_{foot}$ is defined in Equation~\ref{eq:loss_manifold}, penalizing the foot skating artifacts. $L_{phase}$ is the reconstruction loss of the phase, where $F^t$ is calculated as the difference between two consecutive signed shift:
\begin{equation}
 F^t
    \begin{cases}
        S^t-S^{t-1}, & S^t-S^{t-1}\in[-0.5,0.5] \\
        S^t-S^{t-1}+1, & S^t-S^{t-1}\in[-1,-0.5). \\
        S^t-S^{t-1}-1, & S^t-S^{t-1}\in(0.5,1.0]
    \end{cases}
\end{equation}

\section{Implementation}
\paragraph{Data Formatting}We use the 100STYLE motion dataset~\cite{Mason2022Style} and retarget the motions to a skeleton with $23$ joints by removing all the wrist and thumb joints. Further, we subsample the motion sequences to $30$ fps and augment them by mirroring and temporal random cropping~\cite{Jiang_motion_puzzle}. We use 120-frame clips for $M_s$ and 60-frame clips with 20-frame overlapping for the motion manifold. We orient the motions so that first frame faces toward the X-axis~\cite{holden_deep_2016}. However, instead of using features in local space~\cite{Li_2022_GANimator} that is invariant to the global translation and rotation of the motion, our experiments show that representing the starting and target pose in the same coordinate system improves the performance. So we use the global joint orientation by rotations along a forward and up vector $\textbf{r}\in \mathbb{R}^6$~\cite{zhang_mode-adaptive_2018}. Overall, a motion $M\in\mathbb{R}^{23\times 12\times 60}$ includes the global joint position ({$\mathbb{R}^3$}), velocity ({$\mathbb{R}^3$}), and rotation ({$\mathbb{R}^6$}). The style motion $M_s\in\mathbb{R}^{23\times 12\times 120}$ {is chosen at random from a set of data with the same style as $M$.}

\paragraph{Training}
{We chose a 25-frame window at random from the 60-frame clip for one training step} instead of using the whole sequence, which significantly speeds up the training convergence.

We use AmSgrad with parameters ($\beta_1=0.5,\beta_2=0.9$) and a learning rate $1e-3$. To train the sampler, we shuffle all the style sequences randomly. To ensure that our model is robust to different time durations, we randomly sample a sequence whose length is $20$ in the beginning and increases linearly to $40$ in each epoch, similar to \cite{harvey_robust_2020}. We employ AMSGrad with the same setting as before, with a weight decay $1e-4$ for training style encoder to avoid overfitting and $0$ for other modules. More information is available in the supplementary material (SM).

\section{Experiments and Results}
\label{sec:experiments}
We conduct our experiments on a PC with an Nvidia RTX 3090 graphics card, with an Intel I7-11700K CPU and 32G memory. Our method takes on average $1.7$ ms to synthesize one frame.

\paragraph{Data split}100STYLE contains $100$ styles. Since our manifold and sampler can be trained/tested on different datasets, we set up two testing protocols: style-overlapping and style-no-overlapping. We use the last 10 styles as the style-no-overlapping testing data, and use $10\%$ of each style in the first 90 styles as the style-overlapping testing data. The remaining is the training data.


\subsection{Metrics}
We employ three sets of metrics to evaluate the motion quality, the synthesis controllability and the diversity, under various control signals. We first test our model in reconstructing the missing frames of variable lengths ($10$ frames, $20$ frames, and $40$ frames) and measure the reconstruction accuracy. Then we change the time duration and the locations of the target frames (spatial/temporal control), which requires the character to reach the target via different motions. The locations of the target frame and the time duration are changed by: $(x^T)' = x^0 + d\cdot(x^T-x^0)$ and $t_{durations}' = dt\cdot t_{durations}$, where $x^T$ is the location of the target frame projected on the ground, $t_{durations}$ is the time durations of the missing frames. $d$ and $dt$ are the ratio parameters. We set $d=2$ and $d=-1$ to place the target frame before and after the starting frame. We set $dt=2$ and $dt=0.5$ to slow down and speed up the motion.


{We measure controllability using three different metrics. When infilling motion, we use the averaged L2 distance of global joint positions and the Normalized Power Spectrum Similarity (NPSS) in the joint angle space to calculate reconstruction accuracy \cite{harvey_robust_2020}. When the ground truth for spatial/temporal control is unavailable, we measure controllability by determining whether the predicted last frame matches the target. In both cases, we measure foot skating \cite{zhang_mode-adaptive_2018} using the averaged foot velocity $v_f$ when the foot height $h$ is within a threshold $H=2.5$: $L_f = v_f\cdot clamp(2-2^{h/{H}},0,1)$.
}

To evaluate the diversity, we generate ten samples for each combination of a motion sequence and a style sequence, then calculate the L2 distance between the global joint positions of any two sequences. 

Quantitatively measuring style is hard. A possible solution is the Fr\'{e}chet Motion Distance (FMD)~\cite{Jiang_motion_puzzle,yin2023dance}, which compares the distributions of the generated sequences and the dataset. However, the distribution under the spatial/temporal control shifts the distribution significantly. So we only use FMD when the time duration and target location do not change, in which the style similarity is highly related to the distribution similarity. The results of spatial/temporal control are shown in the video.

To compute the FMD, we first train a style classifier by modifying the style encoder in \cite{Jiang_motion_puzzle} by adding a pooling layer and a softmax layer. The pooling layer removes the temporal and the joint axis of the output from the style encoder. Then the softmax layer transforms the latent to the one-hot embedding of the style. The FMD is then calculated between the generated sequences and the ground truth in the latent space after the pooling layer.

\subsection{Manifold Generalization}

\begin{table}[]
\tiny
\centering
\caption{Manifold and Sampler trained on different data: "BA on C" represents the manifold trained on B and the sampler trained on A and the full model tested on C. Note there is no style overlap among A, B, and C. Fuller results are in the SM.}
\label{tab:differentDataset}

\resizebox{1.0\columnwidth}{!}{%
\begin{tabular}{llll}
& \multicolumn{3}{c}{\centering\textbf{L2 norm of global position}} \\ \cline{2-4}
\specialrule{0em}{1pt}{0pt} \multicolumn{1}{l}{Frames} &
  \multicolumn{1}{l}{10} &
  \multicolumn{1}{l}{20} &
  \multicolumn{1}{l}{40} \\ \hline
  \specialrule{0em}{1pt}{1pt}
{AA on A} ({BA on A})      &  \textbf{0.59} (0.64) &  \textbf{0.76} (0.80) &  \textbf{1.31} (1.38) \\ 
{BB on A} ({AB on A})     &  0.78 (0.80) &  1.15 (1.19) &  1.85 (1.92)\\ 
\cline{2-4}
\specialrule{0em}{1pt}{1pt}
{AA on B} ({BA on B})     &  0.63 (0.62) &  0.94 (0.89) &  1.63 (1.53) \\ 
{BB on B} ({AB on B})  & \textbf{0.53} (0.58) &  \textbf{0.68} (0.74) &  \textbf{1.11} (1.23) \\
\cline{2-4}
\specialrule{0em}{1pt}{1pt}
{AA on C} ({BA on C})  &  \textbf{0.80} (0.82) &  1.21 (\textbf{1.18}) &  \textbf{1.95} (1.97) \\ 
{BB on C} ({AB on C})    &  0.97 (0.98) &  1.56 (1.51) &  2.53 (2.44) \\
\specialrule{0em}{1pt}{3pt}
& \multicolumn{3}{c}{\centering\textbf{$100 \times $NPSS }} \\ \cline{2-4}
\specialrule{0em}{1pt}{1pt}
{AA on A} ({BA on A})    & \textbf{0.435} (0.502) & \textbf{1.640} (1.828) & 9.850 (\textbf{9.576})\\
{BB on A} ({AB on A})   & 0.518 (0.534) & 2.195 (2.141) & 13.408 (11.461)\\
\cline{2-4}
\specialrule{0em}{1pt}{1pt}
{AA on B} ({BA on B})     & 0.549 (0.612) & 2.838 (2.395) & 20.981 (18.043) \\ 
{BB on B} ({AB on B})    & \textbf{0.380} (0.504) & \textbf{1.442} (1.922) & \textbf{10.067} (12.387) \\

\cline{2-4}
\specialrule{0em}{1pt}{1pt}
{AA on C} ({BA on C})    & \textbf{0.634} (0.743) & \textbf{3.798} (4.057) & \textbf{16.784} (18.372)\\ 
{BB on C} ({AB on C})   & 0.872 (1.032) & 5.046 (5.541) & 26.806 (19.864) \\
\specialrule{0em}{1pt}{3pt}
& \multicolumn{3}{c}{\centering\textbf{Foot skate}} \\  \cline{2-4}
\specialrule{0em}{1pt}{1pt}
{Ground Truth on A}  &  0.161 &  0.167 &  0.167 \\
{AA on A} ({BA on A})   &  \textbf{0.171} (0.202) &  \textbf{0.197} (0.222) &  \textbf{0.297} (0.321)\\ 
{BB on A} ({AB on A})     &  0.217 (0.199) &  0.250 (0.230) &  0.353 (0.340)\\
\cline{2-4}
\specialrule{0em}{1pt}{1pt}
{Ground Truth on B}   &  0.184 &  0.181 &  0.186 \\
{AA on B} ({BA on B})     &  \textbf{0.211} (0.256) &  \textbf{0.218} (0.232) &  0.324 (0.346)\\
{BB on B} ({AB on B})  &  0.213 (0.221) &  0.231 (0.249) &  \textbf{0.309} (0.343) \\
\cline{2-4}
\specialrule{0em}{1pt}{1pt}
{Ground Truth on C}   &  0.271 &  0.269 &  0.255 \\
{AA on C} ({BA on C})        &  \textbf{0.196} (0.210) &  0.264 (\textbf{0.241}) &  0.336 (\textbf{0.302})\\
{BB on C} ({AB on C}) &  0.235 (0.218) &  0.276 (0.279) &  0.345 (0.358) \\
\specialrule{0em}{1pt}{3pt}
& \multicolumn{3}{c}{\centering\textbf{Diversity}} \\  \cline{2-4}
\specialrule{0em}{1pt}{1pt}
{AA on A} (BA on A)     &  0.869 (0.908) &  2.172 (2.210) &  7.194 (7.423)   \\ 
{BB on A} ({AB on A})    &  0.899 (\textbf{0.938}) &  2.283 (\textbf{ 2.303}) &  7.446 (\textbf{7.587})   \\
\specialrule{0em}{1pt}{0pt}
\cline{2-4}
\specialrule{0em}{1pt}{1pt}
{AA on B} ({BA on B})      &  0.856 (\textbf{0.892}) &  2.189 (\textbf{2.191}) &  7.471 ( \textbf{7.612})  \\
{BB on B} ({AB on B})  &  0.834 (0.875) &  2.049 (2.096) &  6.738 (6.989) \\
\end{tabular}}
\vspace{-4.em}
\end{table}
One advantage of our model is that the manifold and sampler do not have to be trained on the same dataset, which makes it possible to train the manifold on a general dataset with more data samples and train the sampler on a dedicated style dataset. To validate this, we further split the training dataset into two subsets named A (the first $46$ styles) and B (the following $44$ styles) to ensure there is no style overlapping. We name the testing dataset C (the last $10$ styles). 

We show exhaustive experiments in \cref{tab:differentDataset}. Both the reconstruction and the NPSS are better when the sampler is trained and tested on the same dataset, understandably. Similar observation can be made in foot skating except that AA on B is better than AB on B but by a small margin. Overall, these metrics are not severely affected regarding the dataset for manifold training, verifying that manifold, which being high-quality, can be trained as a somewhat independent motion source, which gives great flexibility in training and reduces the need to capture large amounts of dedicated style motions.

In terms of diversity, a higher diversity is achieved in general when the manifold and the sampler are trained on different data. It is challenging to reproduce the original motion when sampling from another manifold, which drives the sampler to approximate it by stochastic sampling, essentially increasing the diversity. Besides, compared with the manifold, the sampler has a stronger influence on diversity, especially when trained and tested on different data. The diversity comes from the fact that the character is forced to explore more poses to reach the target.

FMD (\cref{tab:differentDatasetChangeTimeDistance}) measures the distributional similarity between our model and the ground truth. Again, the sampler has a bigger influence, e.g. both AA and BA are better than AB on A. Further, although BA performs slightly worse than AA on A, the style effects do not differ much in visual observations. It is reasonable because the motion sampled from manifold A has a higher probability of having a similar distribution as dataset A. Similar results (in SM) are also observed on dataset C. 

In terms of the controllability, the metrics on the last frame is mainly affected by the manifold, as shown in \cref{tab:differentDatasetChangeTimeDistance}. The sampling from manifold A achieves better synthesis controllability in dataset A than from manifold B. However, there is no significant visual difference between them in most cases. For some poses that differ from the neutral pose in dataset A, sampling from manifold B cannot reconstruct the pose accurately, resulting in a gap in the target frame. Training the manifold on a more diverse dataset alleviates this problem. Overall, the manifold affects the controllability and mildly influences the quality while the sampler mainly influence the style and the generalization.


\begin{table}[]
\tiny
\centering
\caption{Manifold and Sampler trained on different data. The table shows FMD of $40$ frames and the L2 norm of global position between the last predicted frame and the target frame.}
\label{tab:differentDatasetChangeTimeDistance}

\resizebox{1.0\columnwidth}{!}{%
\begin{tabular}{lllll}
& \multicolumn{4}{c}{\centering\textbf{FMD (40 frames)}} \\  \cline{2-5}
\specialrule{0em}{1pt}{0pt}
Dataset A &{BA } & {AA }   &{AB } &{BB } \\
&16.303    &  \textbf{10.359}     &  50.448 &  58.154  \\
\cline{2-5}
\specialrule{0em}{1pt}{0pt} 
Dataset B &{BA }    &{AA }&{AB } &{BB } \\
&45.898     &  62.092 & 14.964 & \textbf{11.431} \\
\cline{2-5}
\specialrule{0em}{1pt}{0pt} 
Dataset C&{BA }    & {AA }  & {AB } &{BB } \\
&\textbf{70.655}     &  74.404 &   105.360  & 102.639 \\ 
\specialrule{0em}{1pt}{3pt} 
& \multicolumn{4}{c}{\centering\textbf{L2 norm of global position of last frame}}
\\\specialrule{0em}{1pt}{0pt} \cline{2-5}
\specialrule{0em}{1pt}{0pt} \multicolumn{1}{l}{Conditions} &
  \multicolumn{1}{l}{dt=2, d=1} &
  \multicolumn{1}{l}{dt=0.5, d=1} &
  \multicolumn{1}{l}{dt=1, d=2} &
  \multicolumn{1}{l}{dt=1, d=-1}\\ \hline
  \specialrule{0em}{1pt}{1pt}
  
{AA on A} ({AB on A})    &  0.38 (\textbf{0.37}) &  \textbf{0.42} (0.42) &  \textbf{0.39} (0.40) &  0.39 (\textbf{0.37})\\
{BB on A} ({BA on A})      &  0.43 (0.45) &  0.48 (0.48) &  0.46 (0.43) &  0.45 (0.44)\\ 
\cline{2-5}
\specialrule{0em}{1pt}{1pt}
{AA on B} ({AB on B})    &  0.40 (0.37) &  0.43 (0.40) &  0.39 (0.37) &  0.39 (0.36) \\ 
{BB on B} ({BA on B})  &  \textbf{0.33} (0.35) &  \textbf{0.36} (0.38) &  0.33 (\textbf{0.33}) & \textbf{0.32} (0.34) \\ 
\end{tabular}}
\vspace{-2.0em}
\end{table}

\subsection{Comparisons}
We compare our model with two in-between models: CVAE~\cite{tang2022cvae} and RTN~\cite{harvey_robust_2020}. Results are shown in ~\cref{tab:comparison}.
 {Besides, for a fair comparison, an additional experiment was carried out in which we replaced the CVAE's sampler with our sampler so that it could accept the style code as input. The manifold design distinguishes the two methods. The comparison demonstrates that the phase manifold is critical for preserving the styled motion, as discussed in the SM's manifold discussion. Furthermore, we compare our method with and without the attention block, which is also detailed in the SM.}
From reconstruction and NPSS, CVAE performs overall slightly better than ours and both are better than RTN. Since the motion sampling of CVAE is strictly constrained within a latent Gaussian distribution, its NPSS is slightly better than ours while we achieves better reconstruction in general. However, when it comes to foot skating and motion diversity, our method undoubtedly outperforms all baseline methods. We argue that in styled motion generation, the latter two metrics are more important as the foot skating is a key indicator for any motion while the diversity serves the generation task better. 



\begin{figure*}[tb]
	\centering
	\includegraphics[width=0.9\linewidth]{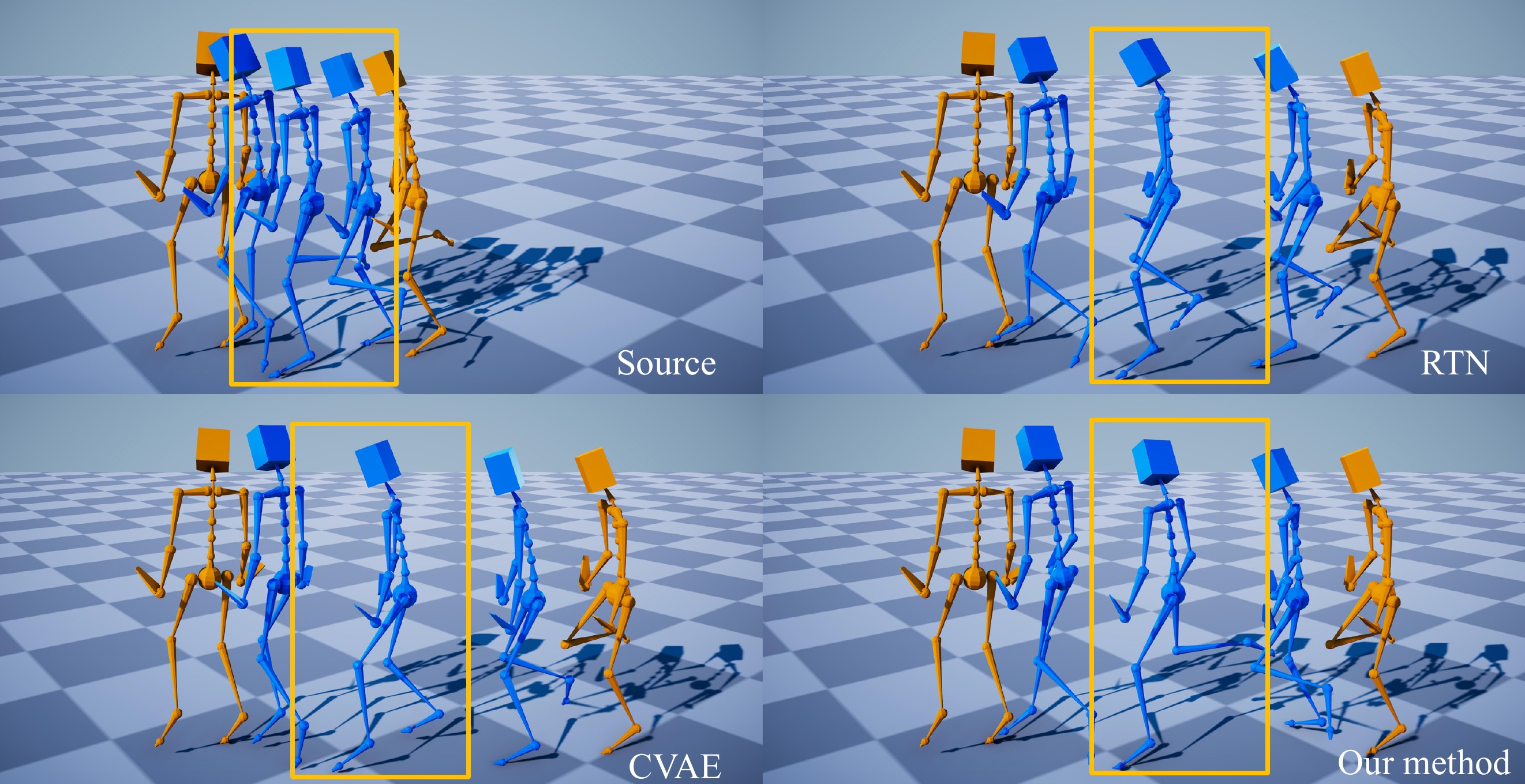}
    \caption{The target location is set to further away in front of the starting frame. Our method performs the desired stylized motion with bigger footsteps. Comparatively, the character synthesized by RTN and CVAE reaches the target via neutral walking. The yellow boxes highlight the most stylized pose. }
	\label{fig:compare_style_flick} 
\end{figure*}

\begin{figure*}[tb]
	\centering
	\includegraphics[width=0.9\linewidth]{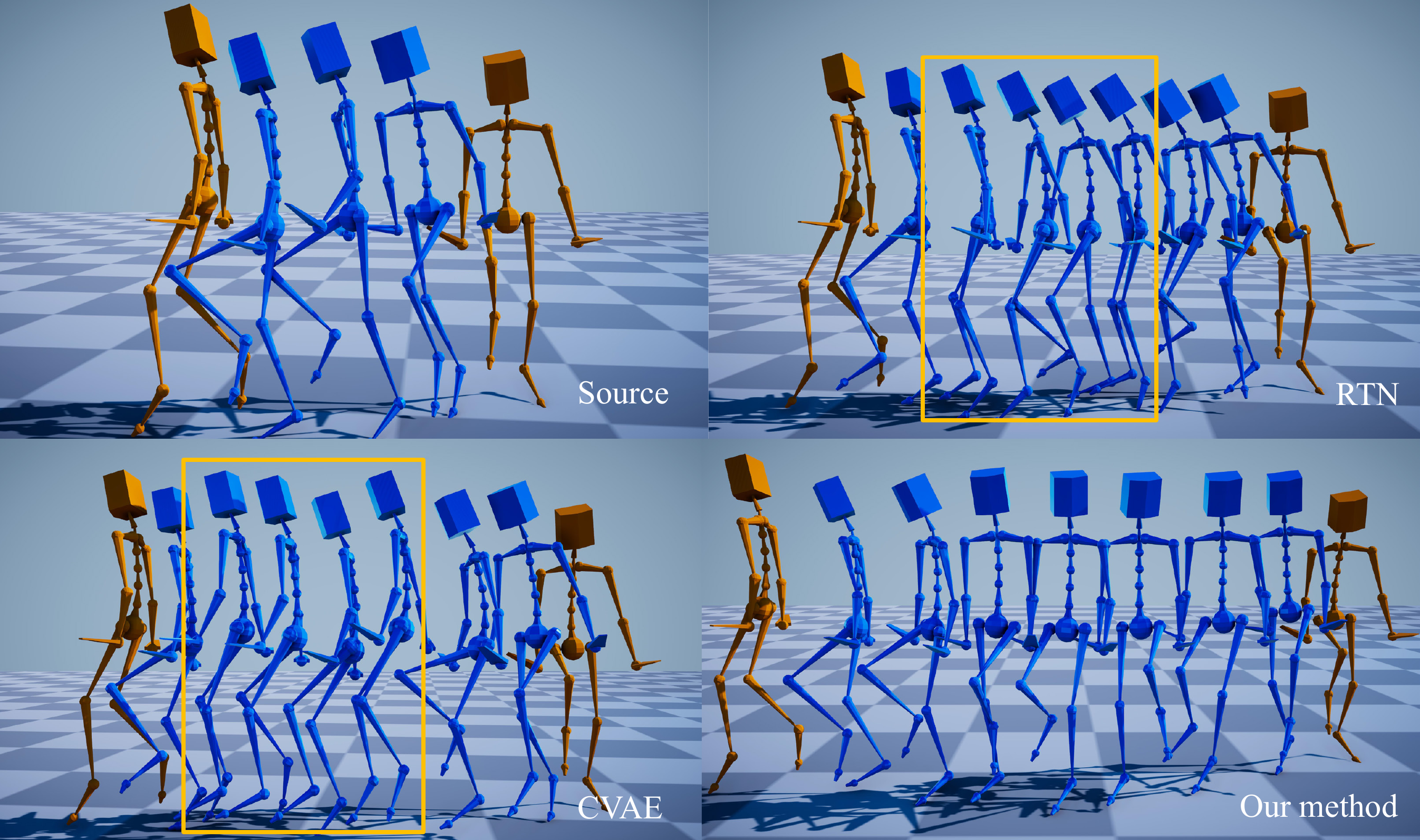}
    \caption{We set the time duration twice as long as the original (i.e. requiring a slowing down motion) and plot the key poses for easy viewing. RTN and CVAE remain in the middle and wait without performing the stylized motion (yellow box), whereas our method continues performing the stylized motion.}
	\label{fig:compare_style_t=2} 
\end{figure*}

Further, without imposing style explicitly, RTN and CVAE simply learn the most likely motion given the past context and the target frame from the datasets. Sometimes they can generate motion with styles but only when there is a style that is dinstinctive from other styles without ambiguity in the data. When there are several similar styles, they tend to mix them incorrectly, which can be noticed in the visual comparison in the video. In addition, when the motion differs from the distribution of the training set by changing the time duration or locations, RTN and CVAE cannot generate vivid styled motions or generate visible artifacts. 
Please refer to Figs. \ref{fig:compare_style_flick}, \ref{fig:compare_style_t=2} in the figures only pages, as well as the accompanying video.
Alternatively, in the absence of style specification, our method achieves similar high-quality results compared to CVAE, and both methods outperform RTN by motion quality, especially for the conditions that the time duration and the locations of the target frames are changed.
\vspace{-0.5em}
\paragraph{Varying transition duration} We also significantly speed up and slow down the velocity. Speeding up dictates that the character needs to make faster steps or larger strides to reach the target. In this situation, CVAE and our method adopt the same phase as the original sequence but with higher velocity. In contrast, RTN drifts to the target. During slowing down, CVAE generates motions where the character stays in the middle and waits without performing the stylized motion, while our method also slows down but still performs the styled motion or keeps moving but turns slightly if it moves to a location that is slightly different from the target, e.g. overshooting. The reason for such motions is that a longer duration drives more cycles in the the phase manifold in our model, so that it avoids simply drifting/idling in the middle like the motions by RTN and CVAE.  More visual comparisons can be found in the video.

\vspace{-0.8em}
\paragraph{Different target locations.} To test drastic spatial control, we run two experiments in which the target location is set to be further away in front of and behind the character, respectively. When the target is further away, the CVAE fills the gap with more small footsteps or fewer bigger ones. Our method uses the bigger steps to fill the gap while preserving the same phase as the original sequence. However, RTN usually performs at the same pace as the ground truth but fills the distance gap by drifting. When the target is behind the character, which is drastically different from the data, the character must change the velocity direction during the motion, causing RTN to easily drift. But our method and CVAE keep foot contact with the ground.

 \begin{table}[]
\tiny
\centering
\caption{Comparison on reconstruction, foot skating, and diversity metrics. }
\label{tab:comparison}

\resizebox{1.0\columnwidth}{!}{%
\begin{tabular}{lllllll}
& \multicolumn{3}{c}{\centering\textbf{L2 norm of global position}} & \multicolumn{3}{c}{\centering\textbf{$100 \times $NPSS}} \\ \cline{2-7}
\specialrule{0em}{1pt}{0pt} \multicolumn{1}{l}{Frames} &
  \multicolumn{1}{l}{10} &
  \multicolumn{1}{l}{20} &
  \multicolumn{1}{l}{40} &
  \multicolumn{1}{l}{10} &
  \multicolumn{1}{l}{20} &
  \multicolumn{1}{l}{40}\\ \hline
  \specialrule{0em}{1pt}{1pt}
{RTN} &  0.663 &  0.847 &  1.365 & 0.387 & 1.539 & 8.794 \\
{CVAE}    &  \textbf{0.506} &  0.692 &  1.224& \textbf{0.321} & \textbf{1.353} & \textbf{8.108} \\ 
{Our method}         &  0.525 &  \textbf{0.680} &  \textbf{1.148} & 0.384 & 1.507 & 9.659 \\
 \specialrule{0em}{1pt}{2pt}
& \multicolumn{3}{c}{\centering\textbf{Foot skate}}& \multicolumn{3}{c}{\centering\textbf{Diversity}} \\  \cline{2-7}
\specialrule{0em}{1pt}{1pt}
{RTN} &  0.343 &  0.339 &  0.474  &  0.678 &  1.956 &  6.172\\
{CVAE}    &  0.217 &  0.210 &  0.284  &  0.867 &  1.791 &  5.560 \\ 
{Our method}         &  \textbf{0.174} &  \textbf{0.194} &  \textbf{0.272} &  \textbf{0.910} &  \textbf{2.017} &  \textbf{6.683 }
\end{tabular}}
\end{table}

\begin{table}[]
\tiny
\centering
\caption{Comparison on the L2 norm of global position of last predicted frame and the target, foot skating metrics, under the conditions that change the time duration and locations of target frame of $40$ missing frames. }
\label{tab:ComparisonChangeTimeDistance}

\resizebox{1.0\columnwidth}{!}{%
\begin{tabular}{lllll}
& \multicolumn{4}{c}{\centering\textbf{L2 norm of global position of the last frame}} \\ \cline{2-5}
\specialrule{0em}{1pt}{0pt} \multicolumn{1}{l}{Conditions} &
  \multicolumn{1}{l}{dt=2, d=1} &
  \multicolumn{1}{l}{dt=0.5, d=1} &
  \multicolumn{1}{l}{dt=1, d=2} &
  \multicolumn{1}{l}{dt=1, d=-1}\\ \hline
\specialrule{0em}{1pt}{0pt}
{RTN}      &  0.688 &  0.599 &  0.751 &  0.599 \\
{CVAE}     &  0.908 &  0.781 &  0.746 &  0.811 \\ 
{Our method}  &  \textbf{0.292} &  \textbf{0.358} &  \textbf{0.302} &  \textbf{0.303}\\ 
& \multicolumn{4}{c}{\centering\textbf{Foot skate}} \\  \cline{2-5}
\specialrule{0em}{1pt}{0pt}
{RTN}      &  0.498 &  1.294 &  1.181 &  0.795 \\
{CVAE}      &  \textbf{0.208} &  \textbf{0.995} &  0.875 &  0.664 \\ 
{Our method}   &  0.307 &  1.018 & \textbf{0.592} &  \textbf{0.557}\\ 
\end{tabular}}
 \vspace{-3.em}
\end{table}

\subsection{Generation on Unseen styles} 
Compared to earlier motion in-between methods~\cite{harvey_robust_2020,tang2022cvae}, which cannot generalize to unseen styles, we have shown our method can generalize well in style-no-overlapping results (\cref{tab:differentDataset} and \cref{tab:differentDatasetChangeTimeDistance}). In addition, the design of our model makes fine-tune on limited data easy, so we do not have to blindly extrapolate on new styles. This is important in application as existing data might not give enough samples in the style space~\cite{Ji2021TestSA} and it is highly desirable if only a few sequences of a new style are needed for models to generate diversified motions with that style.

To validate this, we fine-tune the style encoder and the linear layers of the FiLM and ATN blocks while keeping all other parameters fixed. We choose the "TwoFootJump" style in the testing dataset as it is significantly different from the training set. The "TwoFootJump" data consists of $8$ types. We sample merely one sequence per type and augment the sequences by mirroring and temporal random cropping, leading to $24$ sequences for fine-tuning. After fine-tuning, the character can reach the target by jumping rather than walking as before fine-tuning, which captures the new style perfectly. The accompanying video shows the visual results.





\section{Limitations, Conclusions and Future Work}
One limitation is our model still tends to generate the sequences that have similar distribution as the training data, similar to other data-driven methods. When the specified style severely contradicts with the space-time constraints, our model favors the control rather than the style, especially if the remaining time is not long enough to show that style. For example, the character should kick strongly to deliver the style, but it might raise its leg slightly if the remaining time is insufficient and the target pose does not raise the leg. However, we argue in most application scenarios, control is more important.

In summary, we have proposed a novel learning framework for styled real-time in-between motion generation (RSMT), consisting of two critical, independent components: a general motion manifold acting as a high-quality motion source and a motion sampler generating motions with user-specified style and control signals. Exhaustive evaluation proves that our model has a strong generalization capacity in motion control and stylization. Our method generates high quality styled motions and is general to unseen control signals. It also outperforms state-of-the-art methods. 
One future direction is to enable style transitions before reaching the target frame, which requires tunable weighting on control and varying styles simultaneously on the fly.

\begin{acks}
Xiaogang Jin was supported by Key R\&D Program of Zhejiang (No. 2023C01047), and the National Natural Science Foundation of China (Grant Nos. 62036010, 61972344). He Wang has received funding from the European Union’s Horizon 2020 research and innovation programme under grant agreement No 899739 CrowdDNA.
\end{acks}
\bibliographystyle{ACM-Reference-Format}
\bibliography{sample-base}


\begin{thebibliography}{58}


\ifx \showCODEN    \undefined \def \showCODEN     #1{\unskip}     \fi
\ifx \showDOI      \undefined \def \showDOI       #1{#1}\fi
\ifx \showISBNx    \undefined \def \showISBNx     #1{\unskip}     \fi
\ifx \showISBNxiii \undefined \def \showISBNxiii  #1{\unskip}     \fi
\ifx \showISSN     \undefined \def \showISSN      #1{\unskip}     \fi
\ifx \showLCCN     \undefined \def \showLCCN      #1{\unskip}     \fi
\ifx \shownote     \undefined \def \shownote      #1{#1}          \fi
\ifx \showarticletitle \undefined \def \showarticletitle #1{#1}   \fi
\ifx \showURL      \undefined \def \showURL       {\relax}        \fi
\providecommand\bibfield[2]{#2}
\providecommand\bibinfo[2]{#2}
\providecommand\natexlab[1]{#1}
\providecommand\showeprint[2][]{arXiv:#2}

\bibitem[\protect\citeauthoryear{Aberman, Weng, Lischinski, Cohen-Or, and
  Chen}{Aberman et~al\mbox{.}}{2020}]%
        {aberman2020unpaired}
\bibfield{author}{\bibinfo{person}{Kfir Aberman}, \bibinfo{person}{Yijia Weng},
  \bibinfo{person}{Dani Lischinski}, \bibinfo{person}{Daniel Cohen-Or}, {and}
  \bibinfo{person}{Baoquan Chen}.} \bibinfo{year}{2020}\natexlab{}.
\newblock \showarticletitle{Unpaired motion style transfer from video to
  animation}.
\newblock \bibinfo{journal}{\emph{ACM Transactions on Graphics}}
  \bibinfo{volume}{39}, \bibinfo{number}{4} (\bibinfo{year}{2020}),
  \bibinfo{pages}{1--12}.
\newblock


\bibitem[\protect\citeauthoryear{Arikan and Forsyth}{Arikan and
  Forsyth}{2002}]%
        {arikan_interactive_2002}
\bibfield{author}{\bibinfo{person}{Okan Arikan} {and} \bibinfo{person}{D.~A.
  Forsyth}.} \bibinfo{year}{2002}\natexlab{}.
\newblock \showarticletitle{Interactive motion generation from examples}.
\newblock \bibinfo{journal}{\emph{ACM Transactions on Graphics}}
  \bibinfo{volume}{21}, \bibinfo{number}{3} (\bibinfo{year}{2002}),
  \bibinfo{pages}{483--490}.
\newblock


\bibitem[\protect\citeauthoryear{Beaudoin, Coros, van~de Panne, and
  Poulin}{Beaudoin et~al\mbox{.}}{2008}]%
        {beaudoin2008motion}
\bibfield{author}{\bibinfo{person}{Philippe Beaudoin}, \bibinfo{person}{Stelian
  Coros}, \bibinfo{person}{Michiel van~de Panne}, {and} \bibinfo{person}{Pierre
  Poulin}.} \bibinfo{year}{2008}\natexlab{}.
\newblock \showarticletitle{Motion-motif graphs}. In
  \bibinfo{booktitle}{\emph{Proceedings of the 2008 ACM SIGGRAPH/Eurographics
  Symposium on Computer Animation}}. \bibinfo{pages}{117--126}.
\newblock


\bibitem[\protect\citeauthoryear{Brand and Hertzmann}{Brand and
  Hertzmann}{2000}]%
        {brand2000style}
\bibfield{author}{\bibinfo{person}{Matthew Brand} {and} \bibinfo{person}{Aaron
  Hertzmann}.} \bibinfo{year}{2000}\natexlab{}.
\newblock \showarticletitle{Style machines}. In
  \bibinfo{booktitle}{\emph{Proceedings of the 27th Annual Conference on
  Computer Graphics and Interactive Techniques}}. \bibinfo{pages}{183--192}.
\newblock


\bibitem[\protect\citeauthoryear{Chai and Hodgins}{Chai and Hodgins}{2007}]%
        {chai_constraint-based_2007}
\bibfield{author}{\bibinfo{person}{Jinxiang Chai} {and}
  \bibinfo{person}{Jessica~K. Hodgins}.} \bibinfo{year}{2007}\natexlab{}.
\newblock \showarticletitle{Constraint-based motion optimization using a
  statistical dynamic model}.
\newblock \bibinfo{journal}{\emph{ACM Transactions on Graphics}}
  \bibinfo{volume}{26}, \bibinfo{number}{3} (\bibinfo{year}{2007}),
  \bibinfo{pages}{8--es}.
\newblock


\bibitem[\protect\citeauthoryear{Chen, Wang, Yuan, Shao, and Zhou}{Chen
  et~al\mbox{.}}{2020}]%
        {Chen_dynamic_2020}
\bibfield{author}{\bibinfo{person}{Wenheng Chen}, \bibinfo{person}{He Wang},
  \bibinfo{person}{Yi Yuan}, \bibinfo{person}{Tianjia Shao}, {and}
  \bibinfo{person}{Kun Zhou}.} \bibinfo{year}{2020}\natexlab{}.
\newblock \showarticletitle{Dynamic future net: diversified human motion
  generation}. In \bibinfo{booktitle}{\emph{Proceedings of the 28th ACM
  International Conference on Multimedia}}. \bibinfo{pages}{2131–2139}.
\newblock


\bibitem[\protect\citeauthoryear{Dong, Aristidou, Shamir, Mahler, and
  Jain}{Dong et~al\mbox{.}}{2020}]%
        {dong2020adult2child}
\bibfield{author}{\bibinfo{person}{Yuzhu Dong}, \bibinfo{person}{Andreas
  Aristidou}, \bibinfo{person}{Ariel Shamir}, \bibinfo{person}{Moshe Mahler},
  {and} \bibinfo{person}{Eakta Jain}.} \bibinfo{year}{2020}\natexlab{}.
\newblock \showarticletitle{Adult2child: motion style transfer using
  cyclegans}.
\newblock In \bibinfo{booktitle}{\emph{Proceedings of the 11th Annual
  International Conference on Motion, Interaction, and Games}}.
  \bibinfo{pages}{1--11}.
\newblock


\bibitem[\protect\citeauthoryear{Du, Herrmann, Sprenger, Fischer, and
  Slusallek}{Du et~al\mbox{.}}{2019}]%
        {du2019stylistic}
\bibfield{author}{\bibinfo{person}{Han Du}, \bibinfo{person}{Erik Herrmann},
  \bibinfo{person}{Janis Sprenger}, \bibinfo{person}{Klaus Fischer}, {and}
  \bibinfo{person}{Philipp Slusallek}.} \bibinfo{year}{2019}\natexlab{}.
\newblock \showarticletitle{Stylistic locomotion modeling and synthesis using
  variational generative models}.
\newblock In \bibinfo{booktitle}{\emph{Proceedings of the 11th Annual
  International Conference on Motion, Interaction, and Games}}.
  \bibinfo{pages}{1--10}.
\newblock


\bibitem[\protect\citeauthoryear{Duan, Shi, Zou, Lin, Qian, Zhang, and
  Yuan}{Duan et~al\mbox{.}}{2021}]%
        {duan_single-shot_2021}
\bibfield{author}{\bibinfo{person}{Yinglin Duan}, \bibinfo{person}{Tianyang
  Shi}, \bibinfo{person}{Zhengxia Zou}, \bibinfo{person}{Yenan Lin},
  \bibinfo{person}{Zhehui Qian}, \bibinfo{person}{Bohan Zhang}, {and}
  \bibinfo{person}{Yi Yuan}.} \bibinfo{year}{2021}\natexlab{}.
\newblock \showarticletitle{Single-shot motion completion with transformer}.
\newblock \bibinfo{journal}{\emph{arXiv:2103.00776 [cs]}}
  (\bibinfo{year}{2021}).
\newblock


\bibitem[\protect\citeauthoryear{Harvey and Pal}{Harvey and Pal}{2018}]%
        {harvey2018recurrent}
\bibfield{author}{\bibinfo{person}{F\'{e}lix~G. Harvey} {and}
  \bibinfo{person}{Christopher Pal}.} \bibinfo{year}{2018}\natexlab{}.
\newblock \showarticletitle{Recurrent transition networks for character
  locomotion}. In \bibinfo{booktitle}{\emph{SIGGRAPH Asia 2018 Technical
  Briefs}} \emph{(\bibinfo{series}{SA '18})}. \bibinfo{publisher}{Association
  for Computing Machinery}, \bibinfo{pages}{1--4}.
\newblock


\bibitem[\protect\citeauthoryear{Harvey, Yurick, Nowrouzezahrai, and
  Pal}{Harvey et~al\mbox{.}}{2020}]%
        {harvey_robust_2020}
\bibfield{author}{\bibinfo{person}{Félix~G. Harvey}, \bibinfo{person}{Mike
  Yurick}, \bibinfo{person}{Derek Nowrouzezahrai}, {and}
  \bibinfo{person}{Christopher Pal}.} \bibinfo{year}{2020}\natexlab{}.
\newblock \showarticletitle{Robust motion in-betweening}.
\newblock \bibinfo{journal}{\emph{ACM Transactions on Graphics}}
  \bibinfo{volume}{39}, \bibinfo{number}{4} (\bibinfo{year}{2020}),
  \bibinfo{pages}{1--12}.
\newblock
\showISSN{0730-0301, 1557-7368}


\bibitem[\protect\citeauthoryear{He, Saito, Zachary, Rushmeier, and Zhou}{He
  et~al\mbox{.}}{2022}]%
        {he2022nemf}
\bibfield{author}{\bibinfo{person}{Chengan He}, \bibinfo{person}{Jun Saito},
  \bibinfo{person}{James Zachary}, \bibinfo{person}{Holly Rushmeier}, {and}
  \bibinfo{person}{Yi Zhou}.} \bibinfo{year}{2022}\natexlab{}.
\newblock \showarticletitle{NeMF: Neural Motion Fields for Kinematic
  Animation}. In \bibinfo{booktitle}{\emph{NeurIPS}}.
\newblock


\bibitem[\protect\citeauthoryear{Hernandez, Gall, and Moreno-Noguer}{Hernandez
  et~al\mbox{.}}{2019}]%
        {hernandez2019human}
\bibfield{author}{\bibinfo{person}{Alejandro Hernandez},
  \bibinfo{person}{Jurgen Gall}, {and} \bibinfo{person}{Francesc
  Moreno-Noguer}.} \bibinfo{year}{2019}\natexlab{}.
\newblock \showarticletitle{Human motion prediction via spatio-temporal
  inpainting}. In \bibinfo{booktitle}{\emph{Proceedings of the IEEE/CVF
  International Conference on Computer Vision}}. \bibinfo{pages}{7134--7143}.
\newblock


\bibitem[\protect\citeauthoryear{Holden, Kanoun, Perepichka, and Popa}{Holden
  et~al\mbox{.}}{2020}]%
        {holden_learned_2020}
\bibfield{author}{\bibinfo{person}{Daniel Holden}, \bibinfo{person}{Oussama
  Kanoun}, \bibinfo{person}{Maksym Perepichka}, {and} \bibinfo{person}{Tiberiu
  Popa}.} \bibinfo{year}{2020}\natexlab{}.
\newblock \showarticletitle{Learned motion matching}.
\newblock \bibinfo{journal}{\emph{ACM Transactions on Graphics}}
  \bibinfo{volume}{39}, \bibinfo{number}{4} (\bibinfo{year}{2020}),
  \bibinfo{pages}{1--12}.
\newblock


\bibitem[\protect\citeauthoryear{Holden, Komura, and Saito}{Holden
  et~al\mbox{.}}{2017}]%
        {holden2017phase}
\bibfield{author}{\bibinfo{person}{Daniel Holden}, \bibinfo{person}{Taku
  Komura}, {and} \bibinfo{person}{Jun Saito}.} \bibinfo{year}{2017}\natexlab{}.
\newblock \showarticletitle{Phase-functioned neural networks for character
  control}.
\newblock \bibinfo{journal}{\emph{ACM Transactions on Graphics}}
  \bibinfo{volume}{36}, \bibinfo{number}{4} (\bibinfo{year}{2017}),
  \bibinfo{pages}{1--13}.
\newblock


\bibitem[\protect\citeauthoryear{Holden, Saito, and Komura}{Holden
  et~al\mbox{.}}{2016}]%
        {holden_deep_2016}
\bibfield{author}{\bibinfo{person}{Daniel Holden}, \bibinfo{person}{Jun Saito},
  {and} \bibinfo{person}{Taku Komura}.} \bibinfo{year}{2016}\natexlab{}.
\newblock \showarticletitle{A deep learning framework for character motion
  synthesis and editing}.
\newblock \bibinfo{journal}{\emph{ACM Transactions on Graphics}}
  \bibinfo{volume}{35}, \bibinfo{number}{4} (\bibinfo{year}{2016}),
  \bibinfo{pages}{1--11}.
\newblock


\bibitem[\protect\citeauthoryear{Hsu, Pulli, and Popovi{\'c}}{Hsu
  et~al\mbox{.}}{2005}]%
        {hsu2005style}
\bibfield{author}{\bibinfo{person}{Eugene Hsu}, \bibinfo{person}{Kari Pulli},
  {and} \bibinfo{person}{Jovan Popovi{\'c}}.} \bibinfo{year}{2005}\natexlab{}.
\newblock \showarticletitle{Style translation for human motion}.
\newblock \bibinfo{journal}{\emph{ACM Transactions on Graphics}}
  \bibinfo{volume}{24}, \bibinfo{number}{3} (\bibinfo{year}{2005}),
  \bibinfo{pages}{1082--1089}.
\newblock


\bibitem[\protect\citeauthoryear{Huang and Belongie}{Huang and
  Belongie}{2017}]%
        {huang2017arbitrary}
\bibfield{author}{\bibinfo{person}{Xun Huang} {and} \bibinfo{person}{Serge
  Belongie}.} \bibinfo{year}{2017}\natexlab{}.
\newblock \showarticletitle{Arbitrary style transfer in real-time with adaptive
  instance normalization}. In \bibinfo{booktitle}{\emph{Proceedings of the IEEE
  International Conference on Computer Vision}}. \bibinfo{pages}{1501--1510}.
\newblock


\bibitem[\protect\citeauthoryear{Jang, Park, and Lee}{Jang
  et~al\mbox{.}}{2022}]%
        {Jiang_motion_puzzle}
\bibfield{author}{\bibinfo{person}{Deok-Kyeong Jang}, \bibinfo{person}{Soomin
  Park}, {and} \bibinfo{person}{Sung-Hee Lee}.}
  \bibinfo{year}{2022}\natexlab{}.
\newblock \showarticletitle{Motion puzzle: arbitrary motion style transfer by
  body part}.
\newblock \bibinfo{journal}{\emph{ACM Transactions on Graphics}}
  \bibinfo{volume}{41}, \bibinfo{number}{3} (\bibinfo{year}{2022}),
  \bibinfo{pages}{1--16}.
\newblock


\bibitem[\protect\citeauthoryear{Ji, Pascanu, Hjelm, Lakshminarayanan, and
  Vedaldi}{Ji et~al\mbox{.}}{2021}]%
        {Ji2021TestSA}
\bibfield{author}{\bibinfo{person}{Xu Ji}, \bibinfo{person}{Razvan Pascanu},
  \bibinfo{person}{Devon Hjelm}, \bibinfo{person}{Balaji Lakshminarayanan},
  {and} \bibinfo{person}{Andrea Vedaldi}.} \bibinfo{year}{2021}\natexlab{}.
\newblock \showarticletitle{Test sample accuracy scales with training sample
  density in neural networks}. In \bibinfo{booktitle}{\emph{Proceedings of the
  1st Conference on Lifelong Learning Agents}}. \bibinfo{pages}{629--646}.
\newblock


\bibitem[\protect\citeauthoryear{Kaufmann, Aksan, Song, Pece, Ziegler, and
  Hilliges}{Kaufmann et~al\mbox{.}}{2020}]%
        {kaufmann_convolutional_2020}
\bibfield{author}{\bibinfo{person}{Manuel Kaufmann}, \bibinfo{person}{Emre
  Aksan}, \bibinfo{person}{Jie Song}, \bibinfo{person}{Fabrizio Pece},
  \bibinfo{person}{Remo Ziegler}, {and} \bibinfo{person}{Otmar Hilliges}.}
  \bibinfo{year}{2020}\natexlab{}.
\newblock \showarticletitle{Convolutional autoencoders for human motion
  infilling}. In \bibinfo{booktitle}{\emph{Proceedings of the 2020
  International Conference on 3D Vision}}. \bibinfo{pages}{918--927}.
\newblock


\bibitem[\protect\citeauthoryear{Kingma and Welling}{Kingma and
  Welling}{2013}]%
        {kingma2013auto}
\bibfield{author}{\bibinfo{person}{Diederik~P Kingma} {and}
  \bibinfo{person}{Max Welling}.} \bibinfo{year}{2013}\natexlab{}.
\newblock \showarticletitle{Auto-encoding variational bayes}.
\newblock \bibinfo{journal}{\emph{arXiv preprint arXiv:1312.6114}}
  (\bibinfo{year}{2013}).
\newblock


\bibitem[\protect\citeauthoryear{Kovar, Gleicher, and Pighin}{Kovar
  et~al\mbox{.}}{2008}]%
        {Kovar_motion_2008}
\bibfield{author}{\bibinfo{person}{Lucas Kovar}, \bibinfo{person}{Michael
  Gleicher}, {and} \bibinfo{person}{Fr\'{e}d\'{e}ric Pighin}.}
  \bibinfo{year}{2008}\natexlab{}.
\newblock \showarticletitle{Motion graphs}. In \bibinfo{booktitle}{\emph{ACM
  SIGGRAPH 2008 Classes}} \emph{(\bibinfo{series}{SIGGRAPH '08})}.
  \bibinfo{pages}{1--10}.
\newblock


\bibitem[\protect\citeauthoryear{Levine, Wang, Haraux, Popovi{\'c}, and
  Koltun}{Levine et~al\mbox{.}}{2012}]%
        {levine2012continuous}
\bibfield{author}{\bibinfo{person}{Sergey Levine}, \bibinfo{person}{Jack~M
  Wang}, \bibinfo{person}{Alexis Haraux}, \bibinfo{person}{Zoran Popovi{\'c}},
  {and} \bibinfo{person}{Vladlen Koltun}.} \bibinfo{year}{2012}\natexlab{}.
\newblock \showarticletitle{Continuous character control with low-dimensional
  embeddings}.
\newblock \bibinfo{journal}{\emph{ACM Transactions on Graphics}}
  \bibinfo{volume}{31}, \bibinfo{number}{4} (\bibinfo{year}{2012}),
  \bibinfo{pages}{1--10}.
\newblock


\bibitem[\protect\citeauthoryear{Li, Villegas, Ceylan, Yang, Kuang, Li, and
  Zhao}{Li et~al\mbox{.}}{2021}]%
        {li2021task}
\bibfield{author}{\bibinfo{person}{Jiaman Li}, \bibinfo{person}{Ruben
  Villegas}, \bibinfo{person}{Duygu Ceylan}, \bibinfo{person}{Jimei Yang},
  \bibinfo{person}{Zhengfei Kuang}, \bibinfo{person}{Hao Li}, {and}
  \bibinfo{person}{Yajie Zhao}.} \bibinfo{year}{2021}\natexlab{}.
\newblock \showarticletitle{Task-generic hierarchical human motion prior using
  vaes}. In \bibinfo{booktitle}{\emph{Proceedings of the 2021 International
  Conference on 3D Vision}}. \bibinfo{pages}{771--781}.
\newblock


\bibitem[\protect\citeauthoryear{Li, Aberman, Zhang, Hanocka, and
  Sorkine-Hornung}{Li et~al\mbox{.}}{2022}]%
        {Li_2022_GANimator}
\bibfield{author}{\bibinfo{person}{Peizhuo Li}, \bibinfo{person}{Kfir Aberman},
  \bibinfo{person}{Zihan Zhang}, \bibinfo{person}{Rana Hanocka}, {and}
  \bibinfo{person}{Olga Sorkine-Hornung}.} \bibinfo{year}{2022}\natexlab{}.
\newblock \showarticletitle{GANimator: neural motion synthesis from a single
  sequence}.
\newblock \bibinfo{journal}{\emph{ACM Transactions on Graphics}}
  \bibinfo{volume}{41}, \bibinfo{number}{4} (\bibinfo{year}{2022}),
  \bibinfo{pages}{1--12}.
\newblock


\bibitem[\protect\citeauthoryear{Li, Sun, Zhang, and Wu}{Li
  et~al\mbox{.}}{2013}]%
        {Li_Likelihood_estimation}
\bibfield{author}{\bibinfo{person}{Wanyi Li}, \bibinfo{person}{Jifeng Sun},
  \bibinfo{person}{Xin Zhang}, {and} \bibinfo{person}{Yuanchang Wu}.}
  \bibinfo{year}{2013}\natexlab{}.
\newblock \showarticletitle{Spatial constraints-based maximum likelihood
  estimation for human motions}. In \bibinfo{booktitle}{\emph{Proceedings of
  the 2013 IEEE International Conference on Signal Processing, Communication
  and Computing (ICSPCC 2013)}}. \bibinfo{pages}{1--6}.
\newblock


\bibitem[\protect\citeauthoryear{Li, Wang, and Shum}{Li et~al\mbox{.}}{2002}]%
        {li2002motion}
\bibfield{author}{\bibinfo{person}{Yan Li}, \bibinfo{person}{Tianshu Wang},
  {and} \bibinfo{person}{Heung-Yeung Shum}.} \bibinfo{year}{2002}\natexlab{}.
\newblock \showarticletitle{Motion texture: a two-level statistical model for
  character motion synthesis}. In \bibinfo{booktitle}{\emph{Proceedings of the
  29th Annual Conference on Computer Graphics and Interactive Techniques}}.
  \bibinfo{pages}{465--472}.
\newblock


\bibitem[\protect\citeauthoryear{Ling, Zinno, Cheng, and van~de Panne}{Ling
  et~al\mbox{.}}{2020}]%
        {ling_character_2020}
\bibfield{author}{\bibinfo{person}{Hung~Yu Ling}, \bibinfo{person}{Fabio
  Zinno}, \bibinfo{person}{George Cheng}, {and} \bibinfo{person}{Michiel van~de
  Panne}.} \bibinfo{year}{2020}\natexlab{}.
\newblock \showarticletitle{Character {controllers} {using} {motion} {VAEs}}.
\newblock \bibinfo{journal}{\emph{ACM Transactions on Graphics}}
  \bibinfo{volume}{39}, \bibinfo{number}{4} (\bibinfo{year}{2020}),
  \bibinfo{pages}{1--12}.
\newblock


\bibitem[\protect\citeauthoryear{Ma, Xia, Hodgins, Yang, Li, and Wang}{Ma
  et~al\mbox{.}}{2010}]%
        {ma2010modeling}
\bibfield{author}{\bibinfo{person}{Wanli Ma}, \bibinfo{person}{Shihong Xia},
  \bibinfo{person}{Jessica~K Hodgins}, \bibinfo{person}{Xiao Yang},
  \bibinfo{person}{Chunpeng Li}, {and} \bibinfo{person}{Zhaoqi Wang}.}
  \bibinfo{year}{2010}\natexlab{}.
\newblock \showarticletitle{Modeling style and variation in human motion}. In
  \bibinfo{booktitle}{\emph{Proceedings of the 2010 ACM SIGGRAPH/Eurographics
  Symposium on Computer Animation}}. \bibinfo{pages}{21--30}.
\newblock


\bibitem[\protect\citeauthoryear{Martinez, Black, and Romero}{Martinez
  et~al\mbox{.}}{2017}]%
        {martinez2017human}
\bibfield{author}{\bibinfo{person}{Julieta Martinez},
  \bibinfo{person}{Michael~J Black}, {and} \bibinfo{person}{Javier Romero}.}
  \bibinfo{year}{2017}\natexlab{}.
\newblock \showarticletitle{On human motion prediction using recurrent neural
  networks}. In \bibinfo{booktitle}{\emph{Proceedings of the IEEE Conference on
  Computer Vision and Pattern Recognition}}. \bibinfo{pages}{2891--2900}.
\newblock


\bibitem[\protect\citeauthoryear{Mason, Starke, and Komura}{Mason
  et~al\mbox{.}}{2022}]%
        {Mason2022Style}
\bibfield{author}{\bibinfo{person}{Ian Mason}, \bibinfo{person}{Sebastian
  Starke}, {and} \bibinfo{person}{Taku Komura}.}
  \bibinfo{year}{2022}\natexlab{}.
\newblock \showarticletitle{Real-time style modelling of human locomotion via
  feature-wise transformations and local motion phases}.
\newblock \bibinfo{journal}{\emph{Proceedings of the ACM on Computer Graphics
  and Interactive Techniques}} \bibinfo{volume}{5}, \bibinfo{number}{1}
  (\bibinfo{year}{2022}), \bibinfo{pages}{1--18}.
\newblock


\bibitem[\protect\citeauthoryear{Mason, Starke, Zhang, Bilen, and Komura}{Mason
  et~al\mbox{.}}{2018}]%
        {Mason_2018_fewshot}
\bibfield{author}{\bibinfo{person}{Ian Mason}, \bibinfo{person}{Sebastian
  Starke}, \bibinfo{person}{He Zhang}, \bibinfo{person}{Hakan Bilen}, {and}
  \bibinfo{person}{Taku Komura}.} \bibinfo{year}{2018}\natexlab{}.
\newblock \showarticletitle{Few-shot learning of homogeneous human Locomotion
  Styles}.
\newblock \bibinfo{journal}{\emph{Computer Graphics Forum}}
  \bibinfo{volume}{37}, \bibinfo{number}{7} (\bibinfo{year}{2018}),
  \bibinfo{pages}{143--153}.
\newblock


\bibitem[\protect\citeauthoryear{Min and Chai}{Min and Chai}{2012}]%
        {Min_motion_2012}
\bibfield{author}{\bibinfo{person}{Jianyuan Min} {and}
  \bibinfo{person}{Jinxiang Chai}.} \bibinfo{year}{2012}\natexlab{}.
\newblock \showarticletitle{Motion graphs++: a compact generative model for
  semantic motion analysis and synthesis}.
\newblock \bibinfo{journal}{\emph{ACM Transactions on Graphics}}
  \bibinfo{volume}{31}, \bibinfo{number}{6} (\bibinfo{year}{2012}),
  \bibinfo{pages}{1--12}.
\newblock


\bibitem[\protect\citeauthoryear{Oreshkin, Valkanas, Harvey, M{\'e}nard,
  Bocquelet, and Coates}{Oreshkin et~al\mbox{.}}{2022}]%
        {oreshkin2022delta}
\bibfield{author}{\bibinfo{person}{Boris~N Oreshkin}, \bibinfo{person}{Antonios
  Valkanas}, \bibinfo{person}{F{\'e}lix~G Harvey}, \bibinfo{person}{Louis-Simon
  M{\'e}nard}, \bibinfo{person}{Florent Bocquelet}, {and}
  \bibinfo{person}{Mark~J Coates}.} \bibinfo{year}{2022}\natexlab{}.
\newblock \showarticletitle{Motion inbetweening via deep
  $\Delta$-interpolator}.
\newblock \bibinfo{journal}{\emph{arXiv preprint arXiv:2201.06701}}
  (\bibinfo{year}{2022}).
\newblock


\bibitem[\protect\citeauthoryear{Park, Jang, and Lee}{Park
  et~al\mbox{.}}{2021}]%
        {park2021diverse}
\bibfield{author}{\bibinfo{person}{Soomin Park}, \bibinfo{person}{Deok-Kyeong
  Jang}, {and} \bibinfo{person}{Sung-Hee Lee}.}
  \bibinfo{year}{2021}\natexlab{}.
\newblock \showarticletitle{Diverse motion stylization for multiple style
  domains via spatial-temporal graph-based generative model}.
\newblock \bibinfo{journal}{\emph{Proceedings of the ACM on Computer Graphics
  and Interactive Techniques}} \bibinfo{volume}{4}, \bibinfo{number}{3}
  (\bibinfo{year}{2021}), \bibinfo{pages}{1--17}.
\newblock


\bibitem[\protect\citeauthoryear{Pavlovic, Rehg, and MacCormick}{Pavlovic
  et~al\mbox{.}}{2000}]%
        {pavlovic2000learning}
\bibfield{author}{\bibinfo{person}{Vladimir Pavlovic}, \bibinfo{person}{James~M
  Rehg}, {and} \bibinfo{person}{John MacCormick}.}
  \bibinfo{year}{2000}\natexlab{}.
\newblock \showarticletitle{Learning switching linear models of human motion}.
  In \bibinfo{booktitle}{\emph{Proceedings of the 13th International Conference
  on Neural Information Processing Systems}}. \bibinfo{pages}{942--948}.
\newblock


\bibitem[\protect\citeauthoryear{Petrovich, Black, and Varol}{Petrovich
  et~al\mbox{.}}{2021}]%
        {petrovich2021action}
\bibfield{author}{\bibinfo{person}{Mathis Petrovich},
  \bibinfo{person}{Michael~J Black}, {and} \bibinfo{person}{G{\"u}l Varol}.}
  \bibinfo{year}{2021}\natexlab{}.
\newblock \showarticletitle{Action-conditioned 3d human motion synthesis with
  transformer vae}. In \bibinfo{booktitle}{\emph{Proceedings of the IEEE/CVF
  International Conference on Computer Vision}}. \bibinfo{pages}{10985--10995}.
\newblock


\bibitem[\protect\citeauthoryear{Qin, Zheng, and Zhou}{Qin
  et~al\mbox{.}}{2022}]%
        {qin2022motion}
\bibfield{author}{\bibinfo{person}{Jia Qin}, \bibinfo{person}{Youyi Zheng},
  {and} \bibinfo{person}{Kun Zhou}.} \bibinfo{year}{2022}\natexlab{}.
\newblock \showarticletitle{Motion in-betweening via two-stage transformers}.
\newblock \bibinfo{journal}{\emph{ACM Transactions on Graphics}}
  \bibinfo{volume}{41}, \bibinfo{number}{6} (\bibinfo{year}{2022}),
  \bibinfo{pages}{1--16}.
\newblock


\bibitem[\protect\citeauthoryear{Rempe, Birdal, Hertzmann, Yang, Sridhar, and
  Guibas}{Rempe et~al\mbox{.}}{2021}]%
        {rempe2021humor}
\bibfield{author}{\bibinfo{person}{Davis Rempe}, \bibinfo{person}{Tolga
  Birdal}, \bibinfo{person}{Aaron Hertzmann}, \bibinfo{person}{Jimei Yang},
  \bibinfo{person}{Srinath Sridhar}, {and} \bibinfo{person}{Leonidas~J
  Guibas}.} \bibinfo{year}{2021}\natexlab{}.
\newblock \showarticletitle{Humor: 3d human motion model for robust pose
  estimation}. In \bibinfo{booktitle}{\emph{Proceedings of the IEEE/CVF
  International Conference on Computer Vision}}. \bibinfo{pages}{11488--11499}.
\newblock


\bibitem[\protect\citeauthoryear{Safonova and Hodgins}{Safonova and
  Hodgins}{2007}]%
        {Safonova07constructionand}
\bibfield{author}{\bibinfo{person}{Alla Safonova} {and}
  \bibinfo{person}{Jessica~K. Hodgins}.} \bibinfo{year}{2007}\natexlab{}.
\newblock \showarticletitle{Construction and optimal search of interpolated
  motion graphs}.
\newblock \bibinfo{journal}{\emph{ACM Transactions on Graphics}}
  \bibinfo{volume}{26}, \bibinfo{number}{3} (\bibinfo{year}{2007}),
  \bibinfo{pages}{106--es}.
\newblock


\bibitem[\protect\citeauthoryear{Shen, Wang, Ho, Yang, and Shum}{Shen
  et~al\mbox{.}}{2017}]%
        {shen_posture_2017}
\bibfield{author}{\bibinfo{person}{Yijun Shen}, \bibinfo{person}{He Wang},
  \bibinfo{person}{Edmond S.~L. Ho}, \bibinfo{person}{Longzhi Yang}, {and}
  \bibinfo{person}{Hubert P.~H. Shum}.} \bibinfo{year}{2017}\natexlab{}.
\newblock \showarticletitle{Posture-based and action-based graphs for boxing
  skill visualization}.
\newblock \bibinfo{journal}{\emph{Computers and Graphics}}
  \bibinfo{volume}{69}, \bibinfo{number}{Supplement C} (\bibinfo{year}{2017}),
  \bibinfo{pages}{104--115}.
\newblock


\bibitem[\protect\citeauthoryear{Smith, Cao, Neff, and Wang}{Smith
  et~al\mbox{.}}{2019}]%
        {smith2019efficient}
\bibfield{author}{\bibinfo{person}{Harrison~Jesse Smith}, \bibinfo{person}{Chen
  Cao}, \bibinfo{person}{Michael Neff}, {and} \bibinfo{person}{Yingying Wang}.}
  \bibinfo{year}{2019}\natexlab{}.
\newblock \showarticletitle{Efficient neural networks for real-time motion
  style transfer}.
\newblock \bibinfo{journal}{\emph{Proceedings of the ACM on Computer Graphics
  and Interactive Techniques}} \bibinfo{volume}{2}, \bibinfo{number}{2}
  (\bibinfo{year}{2019}), \bibinfo{pages}{1--17}.
\newblock


\bibitem[\protect\citeauthoryear{Starke, Mason, and Komura}{Starke
  et~al\mbox{.}}{2022}]%
        {Starke_deeppahse_2022}
\bibfield{author}{\bibinfo{person}{Sebastian Starke}, \bibinfo{person}{Ian
  Mason}, {and} \bibinfo{person}{Taku Komura}.}
  \bibinfo{year}{2022}\natexlab{}.
\newblock \showarticletitle{DeepPhase: periodic autoencoders for learning
  motion phase manifolds}.
\newblock \bibinfo{journal}{\emph{ACM Transactions on Graphics}}
  \bibinfo{volume}{41}, \bibinfo{number}{4} (\bibinfo{year}{2022}),
  \bibinfo{pages}{1--13}.
\newblock


\bibitem[\protect\citeauthoryear{Starke, Zhao, Zinno, and Komura}{Starke
  et~al\mbox{.}}{2021}]%
        {starke_neural_2021}
\bibfield{author}{\bibinfo{person}{Sebastian Starke}, \bibinfo{person}{Yiwei
  Zhao}, \bibinfo{person}{Fabio Zinno}, {and} \bibinfo{person}{Taku Komura}.}
  \bibinfo{year}{2021}\natexlab{}.
\newblock \showarticletitle{Neural animation layering for synthesizing martial
  arts movements}.
\newblock \bibinfo{journal}{\emph{ACM Transactions on Graphics}}
  \bibinfo{volume}{40}, \bibinfo{number}{4} (\bibinfo{year}{2021}),
  \bibinfo{pages}{1--16}.
\newblock


\bibitem[\protect\citeauthoryear{Tang, Wang, Hu, Gong, Yi, Kou, and Jin}{Tang
  et~al\mbox{.}}{2022}]%
        {tang2022cvae}
\bibfield{author}{\bibinfo{person}{Xiangjun Tang}, \bibinfo{person}{He Wang},
  \bibinfo{person}{Bo Hu}, \bibinfo{person}{Xu Gong}, \bibinfo{person}{Ruifan
  Yi}, \bibinfo{person}{Qilong Kou}, {and} \bibinfo{person}{Xiaogang Jin}.}
  \bibinfo{year}{2022}\natexlab{}.
\newblock \showarticletitle{Real-time controllable motion transition for
  characters}.
\newblock \bibinfo{journal}{\emph{ACM Transactions on Graphics}}
  \bibinfo{volume}{41}, \bibinfo{number}{4} (\bibinfo{year}{2022}),
  \bibinfo{pages}{1--10}.
\newblock


\bibitem[\protect\citeauthoryear{Tevet, Raab, Gordon, Shafir, Cohen-Or, and
  Bermano}{Tevet et~al\mbox{.}}{2022}]%
        {tevet2022diffusion}
\bibfield{author}{\bibinfo{person}{Guy Tevet}, \bibinfo{person}{Sigal Raab},
  \bibinfo{person}{Brian Gordon}, \bibinfo{person}{Yonatan Shafir},
  \bibinfo{person}{Daniel Cohen-Or}, {and} \bibinfo{person}{Amit~H Bermano}.}
  \bibinfo{year}{2022}\natexlab{}.
\newblock \showarticletitle{Human motion diffusion model}.
\newblock \bibinfo{journal}{\emph{arXiv preprint arXiv:2209.14916}}
  (\bibinfo{year}{2022}).
\newblock


\bibitem[\protect\citeauthoryear{Unuma, Anjyo, and Takeuchi}{Unuma
  et~al\mbox{.}}{1995}]%
        {unuma1995fourier}
\bibfield{author}{\bibinfo{person}{Munetoshi Unuma}, \bibinfo{person}{Ken
  Anjyo}, {and} \bibinfo{person}{Ryozo Takeuchi}.}
  \bibinfo{year}{1995}\natexlab{}.
\newblock \showarticletitle{Fourier principles for emotion-based human figure
  animation}. In \bibinfo{booktitle}{\emph{Proceedings of the 22nd Annual
  Conference on Computer Graphics and Interactive Techniques}}.
  \bibinfo{pages}{91--96}.
\newblock


\bibitem[\protect\citeauthoryear{Wang, Ho, and Komura}{Wang
  et~al\mbox{.}}{2015}]%
        {wang_energy_2015}
\bibfield{author}{\bibinfo{person}{He Wang}, \bibinfo{person}{Edmond~SL Ho},
  {and} \bibinfo{person}{Taku Komura}.} \bibinfo{year}{2015}\natexlab{}.
\newblock \showarticletitle{An energy-driven motion planning method for two
  distant postures}.
\newblock \bibinfo{journal}{\emph{IEEE Transactions on Visualization and
  Computer Graphics}} \bibinfo{volume}{21}, \bibinfo{number}{1}
  (\bibinfo{year}{2015}), \bibinfo{pages}{18--30}.
\newblock


\bibitem[\protect\citeauthoryear{Wang, Ho, Shum, and Zhu}{Wang
  et~al\mbox{.}}{2019}]%
        {Wang_STRNN_2019}
\bibfield{author}{\bibinfo{person}{He Wang}, \bibinfo{person}{Edmond~SL Ho},
  \bibinfo{person}{Hubert~PH Shum}, {and} \bibinfo{person}{Zhanxing Zhu}.}
  \bibinfo{year}{2019}\natexlab{}.
\newblock \showarticletitle{Spatio-temporal manifold learning for human motions
  via long-horizon modeling}.
\newblock \bibinfo{journal}{\emph{IEEE Transactions on Visualization and
  Computer Graphics}} \bibinfo{volume}{27}, \bibinfo{number}{1}
  (\bibinfo{year}{2019}), \bibinfo{pages}{216--227}.
\newblock


\bibitem[\protect\citeauthoryear{Wang and Komura}{Wang and Komura}{2011}]%
        {wang_energy_2011}
\bibfield{author}{\bibinfo{person}{He Wang} {and} \bibinfo{person}{Taku
  Komura}.} \bibinfo{year}{2011}\natexlab{}.
\newblock \showarticletitle{Energy-based pose unfolding and interpolation for
  3D articulated characters}. In \bibinfo{booktitle}{\emph{Proceedings of the
  4th International Conference on Motion in Games}}. \bibinfo{pages}{110--119}.
\newblock


\bibitem[\protect\citeauthoryear{Wang, Sidorov, Sandilands, and Komura}{Wang
  et~al\mbox{.}}{2013}]%
        {wang_harmonic_2013}
\bibfield{author}{\bibinfo{person}{He Wang}, \bibinfo{person}{Kirill~A
  Sidorov}, \bibinfo{person}{Peter Sandilands}, {and} \bibinfo{person}{Taku
  Komura}.} \bibinfo{year}{2013}\natexlab{}.
\newblock \showarticletitle{Harmonic parameterization by electrostatics}.
\newblock \bibinfo{journal}{\emph{ACM Transactions on Graphics}}
  \bibinfo{volume}{32}, \bibinfo{number}{5} (\bibinfo{year}{2013}),
  \bibinfo{pages}{1--12}.
\newblock


\bibitem[\protect\citeauthoryear{Wang, Fleet, and Hertzmann}{Wang
  et~al\mbox{.}}{2007}]%
        {wang2007multifactor}
\bibfield{author}{\bibinfo{person}{Jack~M Wang}, \bibinfo{person}{David~J
  Fleet}, {and} \bibinfo{person}{Aaron Hertzmann}.}
  \bibinfo{year}{2007}\natexlab{}.
\newblock \showarticletitle{Multifactor gaussian process models for
  style-content separation}. In \bibinfo{booktitle}{\emph{Proceedings of the
  24th International Conference on Machine Learning}}.
  \bibinfo{pages}{975--982}.
\newblock


\bibitem[\protect\citeauthoryear{Wen, Yang, Fu, Gao, Sun, and Liu}{Wen
  et~al\mbox{.}}{2021}]%
        {wen2021autoregressive}
\bibfield{author}{\bibinfo{person}{Yu-Hui Wen}, \bibinfo{person}{Zhipeng Yang},
  \bibinfo{person}{Hongbo Fu}, \bibinfo{person}{Lin Gao},
  \bibinfo{person}{Yanan Sun}, {and} \bibinfo{person}{Yong-Jin Liu}.}
  \bibinfo{year}{2021}\natexlab{}.
\newblock \showarticletitle{Autoregressive stylized motion synthesis with
  generative flow}. In \bibinfo{booktitle}{\emph{Proceedings of the IEEE/CVF
  Conference on Computer Vision and Pattern Recognition}}.
  \bibinfo{pages}{13612--13621}.
\newblock


\bibitem[\protect\citeauthoryear{Xia, Wang, Chai, and Hodgins}{Xia
  et~al\mbox{.}}{2015}]%
        {xia2015realtime}
\bibfield{author}{\bibinfo{person}{Shihong Xia}, \bibinfo{person}{Congyi Wang},
  \bibinfo{person}{Jinxiang Chai}, {and} \bibinfo{person}{Jessica Hodgins}.}
  \bibinfo{year}{2015}\natexlab{}.
\newblock \showarticletitle{Realtime style transfer for unlabeled heterogeneous
  human motion}.
\newblock \bibinfo{journal}{\emph{ACM Transactions on Graphics}}
  \bibinfo{volume}{34}, \bibinfo{number}{4} (\bibinfo{year}{2015}),
  \bibinfo{pages}{1--10}.
\newblock


\bibitem[\protect\citeauthoryear{Yin, Yin, Baraka, Kragic, and
  Bj{\"o}rkman}{Yin et~al\mbox{.}}{2023}]%
        {yin2023dance}
\bibfield{author}{\bibinfo{person}{Wenjie Yin}, \bibinfo{person}{Hang Yin},
  \bibinfo{person}{Kim Baraka}, \bibinfo{person}{Danica Kragic}, {and}
  \bibinfo{person}{M{\aa}rten Bj{\"o}rkman}.} \bibinfo{year}{2023}\natexlab{}.
\newblock \showarticletitle{Dance style transfer with cross-modal transformer}.
  In \bibinfo{booktitle}{\emph{Proceedings of the IEEE/CVF Winter Conference on
  Applications of Computer Vision}}. \bibinfo{pages}{5058--5067}.
\newblock


\bibitem[\protect\citeauthoryear{Yumer and Mitra}{Yumer and Mitra}{2016}]%
        {yumer2016spectral}
\bibfield{author}{\bibinfo{person}{M~Ersin Yumer} {and}
  \bibinfo{person}{Niloy~J Mitra}.} \bibinfo{year}{2016}\natexlab{}.
\newblock \showarticletitle{Spectral style transfer for human motion between
  independent actions}.
\newblock \bibinfo{journal}{\emph{ACM Transactions on Graphics}}
  \bibinfo{volume}{35}, \bibinfo{number}{4} (\bibinfo{year}{2016}),
  \bibinfo{pages}{1--8}.
\newblock


\bibitem[\protect\citeauthoryear{Zhang, Starke, Komura, and Saito}{Zhang
  et~al\mbox{.}}{2018}]%
        {zhang_mode-adaptive_2018}
\bibfield{author}{\bibinfo{person}{He Zhang}, \bibinfo{person}{Sebastian
  Starke}, \bibinfo{person}{Taku Komura}, {and} \bibinfo{person}{Jun Saito}.}
  \bibinfo{year}{2018}\natexlab{}.
\newblock \showarticletitle{Mode-adaptive neural networks for quadruped motion
  control}.
\newblock \bibinfo{journal}{\emph{ACM Transactions on Graphics}}
  \bibinfo{volume}{37}, \bibinfo{number}{4} (\bibinfo{year}{2018}),
  \bibinfo{pages}{1--11}.
\newblock


\end{thebibliography}

\clearpage

\end{document}